\documentclass{article}
\usepackage[numbers]{natbib}
\usepackage[final]{neurips_2024}

\usepackage[utf8]{inputenc}
\usepackage[T1]{fontenc}    
\usepackage{url}          
\usepackage{booktabs}       
\usepackage{amsfonts}      
\usepackage{nicefrac}       
\usepackage{microtype}      
\usepackage{xcolor}         
\usepackage{graphicx}
\usepackage{amsfonts}      
\usepackage{nicefrac}
\usepackage[english]{babel} 
\usepackage{amsthm}
\usepackage{amssymb}
\usepackage{mathrsfs}
\usepackage{amsfonts}
\usepackage{amsmath}
\usepackage{multirow}
\usepackage{mathrsfs}
\usepackage{dsfont}
\usepackage{subcaption}
\usepackage{wrapfig}
\usepackage{colortbl}

\newtheorem{theorem}{Theorem}
\newtheorem{prop}{Proposition}

\newtheorem{definition}{Definition}
\newtheorem{assumption}[theorem]{Assumption}

\definecolor{darkblue}{rgb}{0,0.08,0.45}
\usepackage[pagebackref=true,unicode=true,colorlinks=true,citecolor=darkblue,linkcolor=darkblue,urlcolor=darkblue]{hyperref}

\newcommand{\cF}{\mathcal{F}}

\newcommand{\cL}{\mathcal{L}}

\newcommand{\cX}{\mathcal{X}}
\newcommand{\cXY}{{\mathcal{X}\times\mathcal{Y}}}
\newcommand{\cY}{\mathcal{Y}}
\newcommand{\cYk}{\mathcal{Y}_k}

\newcommand{\bE}{\mathbb{E}}
\newcommand{\bI}{\mathds{1}}
\newcommand{\bN}{\mathbb{N}}

\newcommand{\bR}{\mathbb{R}}

\newcommand{\mF}{\mathscr{F}}

\newcommand{\argmin}{\mathrm{argmin}}

\newcommand{\dx}{\mathrm{d}x}
\newcommand{\dy}{\mathrm{d}y}

\newcommand{\pxy}{p(y \mid x)}

\newcommand{\nhyp}{n}
\newcommand{\ntar}{m}
\newcommand{\bn}{[\![1,\nhyp]\!]}
\newcommand{\bm}{[\![1,\ntar]\!]}

\title{Annealed Multiple Choice Learning: Overcoming limitations of Winner-takes-all with annealing
}

\author{
  David Perera$^{*\,1}$\\
  \texttt{david.perera@telecom-paris.fr}
  \And
  Victor Letzelter$^{*\,1\,2}$\\ 
  \texttt{victor.letzelter@telecom-paris.fr}
  \And
  Théo Mariotte$^{1}$\\
  \And
  Adrien Cortés$^{3}$\\
  \And
  Mickael Chen$^{2}$\\
  \And 
  Slim Essid$^{1}$\\
  \And
  Gaël Richard$^{1}$\\
   \AND\normalfont $^1$ LTCI, Télécom Paris, Institut Polytechnique de Paris \\ \normalfont $^2$ Valeo.ai\\ \normalfont $^3$ Sorbonne Université
}

\begin{document}

\maketitle
\def\thefootnote{*}\footnotetext{Equal Contribution.}\def\thefootnote{\arabic{footnote}}

\begin{abstract}
  We introduce Annealed Multiple Choice Learning (aMCL) which combines simulated annealing with MCL. MCL is a learning framework handling ambiguous tasks by predicting a small set of plausible hypotheses. These hypotheses are trained using the Winner-takes-all (WTA) scheme, which promotes the diversity of the predictions. However, this scheme may converge toward an arbitrarily suboptimal local minimum, due to the greedy nature of WTA. We overcome this limitation using annealing, which enhances the exploration of the hypothesis space during training. We leverage insights from statistical physics and information theory to provide a detailed description of the model training trajectory. Additionally, we validate our algorithm by extensive experiments on synthetic datasets, on the standard UCI benchmark, and on speech separation.
\end{abstract}

\section{Introduction}

Ambiguous prediction tasks arise in deep learning when the target $y$ is ill-defined from the input $x$. 
Predicting $y$ directly from $x$ can be challenging due to the partial predictability of $y$ from the information contained in $x$.
Multiple Choice Learning (MCL) \cite{guzman2012multiple, lee2016stochastic} addresses these challenges by providing a small set of plausible \textit{hypotheses}, each representing a different possible outcome given the input. MCL learns these hypotheses using a competitive training scheme that promotes the specialization of the hypotheses in distinct regions of the output space $\cY$. The framework iteratively partitions $\cY$ into a Voronoi tesselation and guides each hypothesis toward the barycenter of its respective Voronoi cell \cite{rupprecht2017learning}. This mechanism makes MCL akin to a gradient-descent-based and conditional variant of the popular K-means algorithm \cite{lloyd1982least}. Like K-means, MCL is sensitive to initialization, subject to hypothesis collapse \cite{makansi2019overcoming}, and more generally may converge toward arbitrarily suboptimal hypothesis configurations \cite{rupprecht2017learning}. While there is a substantial body of literature addressing the limitations of K-means \cite{arthur2007k,franti2019much,pena1999empirical}, relatively little has been done to address these challenges in the context of MCL \cite{rupprecht2017learning, makansi2019overcoming, narayanan2021divide}. The core issue of MCL lies with its greedy gradient-based update of the hypotheses. This greediness precludes the exploration of the hypothesis space, preventing MCL from optimally capturing the ambiguity of $y$. We propose to incorporate annealing into this gradient descent update in order to improve the robustness of MCL. 

Simulated annealing, inspired by the gradual cooling of metals, was originally introduced 
for statistical mechanics applications \cite{hastings1970monte,metropolis1953equation} and was later applied to combinatorial problems \cite{kirkpatrick1983optimization}. It is a random exploration process concurrent to the popular stochastic gradient descent, with a significant difference: gradient descent always tries to improve performance while annealing also accepts to temporarily degrade it for the sake of exploration. The range of this exploration is controlled by a temperature parameter: with a high temperature, annealing explores wide regions of the search space; when the temperature decreases, the exploration becomes narrow, and the system is able to refine its performance. 
This strategy has been shown to converge to an optimal state, provided that the cooling is sufficiently slow \cite{hajek1988cooling}. 

Deterministic annealing \cite{rose1992vector} is a variant of simulated annealing. In simulated annealing, exploration relies on a sequence of random moves across the search space, whereas deterministic annealing seeks greater efficiency by replacing this random process with the exact minimization of a deterministic functional, namely the free energy of the system. It has been shown that deterministic annealing can be efficiently applied to clustering \cite{rose1990statistical, rose1992vector}. In this article, we show that it can be further adapted to the conditional and gradient-based setting of MCL. The resulting algorithm, which we name aMCL for \emph{annealed MCL}, addresses the main issues of MCL and achieves strong performance in practical settings while being straightforward to implement and amenable to analysis. Specifically, we make the following contributions.

\textbf{We introduce Annealed Multiple Choice Learning (aMCL)}, a novel algorithm that incorporates annealing into the multiple choice learning framework (Section \ref{sec:method}).

\textbf{We propose a theoretical analysis of aMCL}, to understand its advantages in comparison to vanilla MCL. We characterize the training trajectory of the model, by establishing an analogy with statistical physics and information theory (Section \ref{sec:theory}).

\textbf{We provide extensive experimental validation,} by applying this method i) to illustrative synthetic examples; ii) to a standard distribution estimation benchmark (UCI datasets); and also iii) to the challenging audio task of speech separation (Section \ref{sec:experiments}). The accompanying code is made available.\footnote{\url{https://github.com/Victorletzelter/annealed_mcl}}

\section{Related Work}

\textbf{Multiple choice learning.} MCL has been successfully applied to various machine learning tasks, typically using multi-head neural networks, with each head providing a prediction \cite{lee2017confident, garcia2021distillation, letzelter2023resilient}. Several works observed the phenomenon of \textit{hypothesis collapse} \cite{brodie2018alpha,firman2018diversenet,ilg2018uncertainty,lee2017confident,tian2019versatile,rupprecht2017learning}, where some hypotheses are left unused during training. Various solutions have been proposed to tackle collapse \cite{rupprecht2017learning, makansi2019overcoming, narayanan2021divide}. Notably, \cite{rupprecht2017learning} introduces Relaxed-WTA, which updates non-winning hypotheses with a gradient scaled by a small constant $\varepsilon$. However, this small gradient biases the hypotheses toward the global barycenter of the target distribution, which can be shown to be suboptimal \cite{du1999centroidal}.

\textbf{Simulated and deterministic annealing.} Deterministic annealing is a variant of simulated annealing \cite{hastings1970monte,metropolis1953equation,kirkpatrick1983optimization}. Rose \textit{et al.} extensively investigated its properties, particularly in relation to statistical physics and clustering \cite{rose1990statistical, rose1992vector, rose1994mapping, rose1998deterministic}. We are, to the best of our knowledge, the first to combine this technique with Winner-takes-all training in a conditional setting. 
    
\textbf{Information theory and quantization.} Quantization \cite{shannon1959coding} consists of discretizing continuous variables over a finite set of symbols. The rate-distortion theory studies the minimal number of bits necessary to encode information at a given level of quantization error \cite{blahut1972computation, arimoto1972algorithm, berger2003rate}. Recently, a relation has been established between optimal quantization of conditional distributions \cite{du1999centroidal} and multiple choice learning \cite{rupprecht2017learning, letzelter2023resilient, letzelter24winner}. In this paper, we propose 
to integrate annealing for conditional quantization.

\section{Annealed Multiple Choice Learning}\label{sec:method}

\subsection{Winner-takes-all loss and its limitations}\label{sec:mcl}

Let $\cX$ and $\cY$ denote subsets of Euclidean vector spaces. We are interested in so-called \textit{ambiguous tasks}, \textit{i.e.}, for any given input $x\in\cX$, there may be several plausible outputs $y\in\cY$. Formally, let $p(x,y)$ denote a joint distribution on $\cX\times\cY$. Multiple Choice Learning (MCL) \cite{guzman2012multiple, lee2016stochastic} was proposed to train neural networks in this setting, and has proven its effectiveness in a wide range of machine vision \cite{lee2017confident}, natural language \cite{garcia2021distillation} and signal processing tasks \cite{letzelter2023resilient}.

MCL consists in training several predictors $(f_1,\dots,f_n) \in \mathcal{F}(\mathcal{X},\mathcal{Y}^n)$, typically a multi-head neural network derived from a common backbone, such that for each input $x \in \mathcal{X}$, the predictions $(f_{1}(x), \dots, f_{n}(x))$ provide an efficient \textit{quantization} of the conditional distribution $p(y \mid x)$ \cite{rupprecht2017learning, letzelter2023resilient, letzelter24winner}. This goal is achieved by minimizing the \textit{distortion} 
 \begin{equation}
  D(f) \triangleq \int_{\cX \times \cY} \min_{k \in \bn} \ell\left(y, f_k(x)\right) p(x,y) \dx \dy\;,
 \label{eq:distortion}
 \end{equation} where $\ell: \mathcal{Y}^2 \rightarrow \mathbb{R}$ is an underlying loss function, for instance, the squared Euclidean distance $\ell(\hat{y},y) = \lVert y - \hat{y} \rVert^2$. Eq. \eqref{eq:distortion} can be seen as a generalization of the conditional distortion \cite{pages2003optimal}.

More specifically, MCL training is an iterative procedure optimizing \eqref{eq:distortion} by alternating the two following steps.
\begin{enumerate}
\item Assign each $y$ to the closest hypothesis $f_k(x)$ to build the Voronoi cells:
\begin{equation}\label{eq:wta_assignation}
\cYk(x)\triangleq\left\{y\in \mathcal{Y} \;|\; \forall l \in \bn,\; \ell(y,f_k(x))\leq \ell(y,f_l(x))\right\}\;.  \end{equation}
\item Minimize the distortion within each cell by taking a gradient step on the WTA loss:
\begin{equation}\label{eq:wta_risk}
\mathcal{L}^{\mathrm{WTA}}(f) \triangleq \int_\cX \sum_{k=1}^{n} \left( \int_{\cY_k(x)} \ell(f_k(x),y) p(y \mid x) \dy \right) p(x) \dx\;.
\end{equation}
\end{enumerate}

The prediction models can be paired with scoring models $(\gamma_1,\dots,\gamma_n) \in \mathcal{F}(\cX,[0,1]^n)$, which are trained to estimate the Voronoi regions' probability mass using the scoring loss \cite{letzelter2023resilient}
\begin{equation}
\mathcal{L}^{\mathrm{scoring}}(\gamma) \triangleq \int_{\mathcal{X} \times \mathcal{Y}} \sum_{k=1}^\nhyp\mathrm{BCE}
\left(\mathds{1}\left[y\in\mathcal{Y}_k(x)\right],\gamma_k(x)\right) p(x,y) \mathrm{d}x \mathrm{d}y\,,
\label{eq:scoring_risk}
\end{equation}
where $\mathrm{BCE}(p, q) \triangleq - p\log(q) - (1-p)\log(1-q)$.
In practice, the two losses \eqref{eq:wta_risk} and \eqref{eq:scoring_risk} are optimized in a compound objective $\mathcal{L} = \mathcal{L}^{\mathrm{WTA}} + \mathcal{L}^{\mathrm{scoring}}$. A probabilistic interpretation of such trained predictors has been developed \cite{rupprecht2017learning, letzelter2023resilient}. It shows that the predictions and scores can be interpreted as a mixture model approximating the conditional density $p(y \mid x)$ by $\sum_{k=1}^{n} \gamma_k(x) \delta_{f_k(x)}(y).$

It has been shown that WTA is sensitive to initialization \cite{narayanan2021divide}, and often leads to suboptimal hypothesis positions, similarly to K-means.  Indeed, WTA is a greedy procedure that updates only the best hypotheses: if one hypothesis falls outside the support of the data density $p(\cdot \mid x)$, it may be isolated from its competitors at initialization, and remain so across training (the \textit{collapse} issue).   

\begin{figure}
    \centering
    \includegraphics[width=0.8\columnwidth]{
    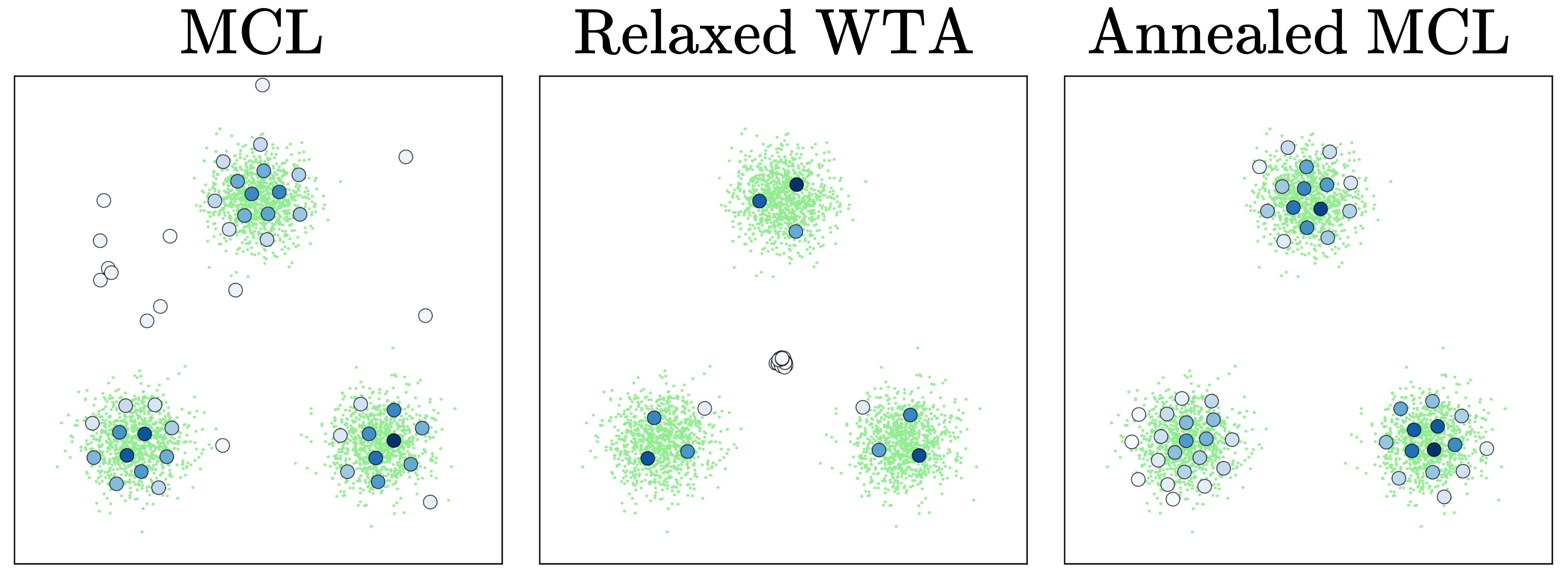
    }
    \caption{\textbf{Overcoming limitations of Winner-Takes-All training with annealing}. Illustrations of the test-time predictions on a Mixture of three Gaussians (green points) with $49$ hypotheses. Shaded blue circles represent the hypothesis predictions, with intensity corresponding to the predicted scores. \textit{(Left)} Predictions of MCL as proposed in \cite{lee2016stochastic, letzelter2023resilient}. \textit{(Middle)} Predictions of Relaxed WTA \cite{rupprecht2017learning} with $\varepsilon = 0.1$. \textit{(Right)} Annealed MCL with initial temperature $T_{0} = 0.6$. Each model was trained with the same backbone (a three-layer MLP). We see that WTA leaves out some hypotheses, achieving a higher quantization error than aMCL. Moreover, we see that Relaxed-WTA is biased toward the barycenter of the distribution, in contrast with aMCL.}
    \label{fig:amcl_comparison}
\end{figure}

Our method improves the WTA training scheme by addressing the inherent greediness of gradient descent and introducing variability in the exploration of the hypothesis space through deterministic annealing.
Figure \ref{fig:amcl_comparison} illustrates the limitations of the aforementioned algorithms and the comparative advantage of aMCL. 

\subsection{Combining deterministic annealing with Multiple Choice Learning}\label{sec:amcl}

We introduce aMCL, which combines MCL and annealing. Let $t\mapsto T(t)$ denote a temperature schedule decreasing with the training step $t$, and vanishing at the end of the training. Similarly to MCL, aMCL alternates between an assignation and a minimization step, as follows:

\begin{enumerate}
    \item Softly assign each $y$ to all $f_k(x)$ using the $\mathrm{softmin}$ operator (or Boltzman distribution $q_{T(t)}$):
    \begin{equation}q_{T(t)}(f_k \mid x, y) \triangleq \frac{1}{Z_{x,y}} \exp \left(-\frac{\ell(f_k(x),y)}{T(t)}\right), \;\;\;  Z_{x,y} \triangleq \sum_{s = 1}^{n} \exp \left(-\frac{\ell(f_s(x),y)}{T(t)} \right),\label{eq:awta_assignation}
    \end{equation}
    \item Minimize the distortion within each soft cell by taking a gradient step on the aWTA loss: 
    \begin{equation}
    \mathcal{L}_{T(t)}^{\mathrm{aWTA}}(f) \triangleq \int_{\mathcal{X} \times \mathcal{Y}} \sum_{k=1}^{n} \ell(f_{k}(x),y) q_{T(t)}(f_k \mid x, y) p(x,y) \mathrm{d}x \mathrm{d}y\;,
    \label{eq:awta_loss}
    \end{equation} 
    where $q_{T(t)}$ is kept constant (\textit{i.e.,} the $\mathrm{stop\_gradient}$ operator is applied).
\end{enumerate}

Therefore, at the lowest level, aMCL simply consists of replacing the $\mathrm{min}$ operator from \eqref{eq:wta_assignation} by $\mathrm{softmin}$. aMCL introduces the temperature schedule as an additional hyperparameter. As highlighted by the literature on simulated annealing \cite{hajek1988cooling}, it is crucial to ensure that the temperature decreases slowly enough to benefit from the advantages of annealing. In practice, we experimented with both linear and exponential schedulers (see also Section \ref{sec:experiments}).

On a higher level, we can interpret the objective of aMCL as a smoothed version of the MCL objective. Smoothing with high temperature simplifies the optimization problem \eqref{eq:awta_loss}, making the loss landscape easier to navigate: we can conjecture from this analysis that aMCL will find a global minimum at high temperature, and we can expect it to stay optimal as long as the temperature decreases slowly enough \cite{csiszar1974computation}. We can also see aMCL as an input-dependent version of deterministic annealing \cite{rose1990statistical, rose1992vector}. In this view, a high temperature encourages the exploration of the hypothesis space and mitigates the greediness of the gradient descent update \eqref{eq:wta_risk}. Moreover, following \cite{hajek1988cooling}, we can posit that there exists an optimal temperature schedule striking a balance between exploration and optimization. Yet another interpretation is that aMCL constitutes an adaptative extension of Relaxed-MCL \cite{rupprecht2017learning}, as $q_{T(t)}(f_k\;|\;x,y)$ depends both on the distance between the hypothesis $f_k(x)$ and the target $y$, and the training step $t$. These interpretations shed light on the inner workings of aMCL.
However, the complete training dynamic of the algorithm appears when we analyze aMCL through the lens of information theory and statistical physics, which is the purpose of the next section.

\section{Theoretical analysis}
\label{sec:theory}

In this Section, we theoretically investigate the properties of our algorithm. Specifically, we detail its training dynamic in Section \ref{sec:alternating_optimization}. In Section \ref{sec:rate_distortion}, we explore how this dynamic relates to the rate-distortion curve. This relationship allows us to study in Section \ref{sec:phase_transition} the phenomenon of \textit{phase transition}, where hypotheses merge and split into sub-groups depending on the temperature. Throughout this Section, we will focus on the squared Euclidean distance $\ell(\hat{y},y) = \lVert y - \hat{y} \rVert^2$. 

\subsection{Soft assignation and entropy constraints}
\label{sec:alternating_optimization}

Minimizing \eqref{eq:distortion} is NP-hard \cite{aloise2009np,dasgupta2008hardness,mahajan2012planar}. Unsurprisingly, MCL can get trapped in local minima during training. In this Section, we discuss why the aMCL training scheme in Section \ref{sec:amcl} is more resilient to this pitfall. The first step toward our analysis is to observe that the $\mathrm{stop\_gradient}$ operator used in \eqref{eq:awta_loss} of the aMCL update effectively turns the algorithm into an alternating optimization of the soft distortion
\begin{equation}
    D(q,f)\triangleq\int_\cXY\sum_{k=1}^n \ell(f_k(x), y) \;q(f_k | x, y)\; p(x, y) \dx\dy\;,
    \label{eq:soft_distortion}
\end{equation} 
where the variables $q$ and $f$ are treated as independent, a procedure similar to Expectation Maximization \cite{dempster1977maximum}. This observation is captured by Proposition \ref{prop:alternating_opt}, where $\cF=\cF(\cX,\cY^n)$ denotes the set of functions from $\cX$ to $\cY^n$, $\Delta_n$ the set of all distributions on $n$ items conditioned by points on $\cX\times\cY$, $H(\pi)=-\sum_{k=1}^n \pi_k\log \pi_k$ the entropy of a discrete distribution $\pi$, $H_T=H(q_T)$ the entropy of the Boltzmann distribution at temperature $T$, and $\lambda_t$ the learning rate of the gradient descent at step $t$.

\begin{prop}[Entropy-constrained alternated minimization] The assignation \eqref{eq:awta_assignation} and optimization \eqref{eq:awta_loss} steps of aMCL correspond to an entropy-constrained block coordinate descent on the soft distortion.

\begin{equation}
    q \leftarrow \underset{{\substack{q \in \Delta_n \\ H(q) \geq H_{T(t)}}}}{\mathrm{argmin}} D(q, f)\;, \qquad  
    f_k \leftarrow f_k - \lambda_t \nabla_{f_k} D(q, f)\;, \;\;\;\forall k \in \bn.
\end{equation}
\label{prop:alternating_opt}
\end{prop}

This is a corollary of Proposition \ref{prop:amcl_training_dynamic}, which provides additional insights on the training dynamics of aMCL.

\begin{prop}[aMCL training dynamic] \label{prop:amcl_training_dynamic} See Proposition \ref{propapp:awta_dynamic} in Appendix.
The following statements are true for all $T>0$, $f\in\cF$, strictly positive $q\in\Delta_n$, $x\in\cX$ and $y\in\cY$. 
\begin{alignat*}{3}
    &(i) \quad  \underset{{\substack{q\in\Delta_n \\ H(q) \geq H_T}}}{\mathrm{argmin}} D(q,f) =q_T\;, \quad && q_T(f_k | x, y) = \frac{\exp\left(-\ell(f_k(x),y\right)/T)}{\sum_{s=1}^n \exp\left(-\ell(f_s(x),y\right)/T)}\;, \quad && \forall k \in \bn \\
    &(ii) \quad  \argmin_{f\in\cF} D(q,f)= f^{\star}\;, \quad && f^{\star}_k(x)=\frac{\int_\cY y \; q(f_k^{\star} | x, y)p(y \mid x)\dy}{\int_\cY q(f_k^{\star} | x, y)p(y \mid x)\dy}\;, \quad && \forall k \in \bn \\
   & (iii)  \quad  \nabla_{f_k} D(q,f) = \gamma^{\star}_k (f_k-f^{\star}_k)\;, \;\;\;\; && \gamma^{\star}_k= \int_\cY q(f_k^{\star} | x, y)p(y \mid x)\dy\;, \quad && \forall k \in \bn
\end{alignat*}

\end{prop}
Part $(i)$ states that the $\mathrm{softmin}$ operator is the solution of the entropy-constrained minimization of the soft distortion. Part $(ii)$ states that a necessary condition for minimizing the soft distortion is that each $f_k$ is a soft barycenter of the assignation distribution for each temperature $T$. Part $(iii)$ states that each gradient update moves $f_k$ toward this soft barycenter $f^{\star}$, and that the update speed depends on the probability mass $\gamma^{\star}_k$ of the points softly assigned to $f_k$. Together, they describe the training dynamics of aMCL. Note that as $T \rightarrow 0$, $q_T$ converges to a one-hot vector, and the soft barycenter in $(ii)$ becomes a hard barycenter. This is consistent with the necessary optimal condition for MCL, $f_k^{\star}(x) = \mathbb{E}_{Y \sim p(y \mid x)}[Y\;|\;Y \in \mathcal{Y}_k(x)]$, proved by Rupperecht \textit{et al.} \cite{rupprecht2017learning}. 
 
\subsection{Rate-distortion curve}
\label{sec:rate_distortion}

We have established in Section \ref{sec:alternating_optimization} that the aMCL training scheme is equivalent to an entropy-constrained alternating optimization of the soft-distortion \eqref{eq:soft_distortion}, with each hypothesis $f_k$ moving toward a soft barycenter. In this Section, we describe the impact of temperature cooling on this training dynamic.

\begin{figure}[!b]
    \centering
    \includegraphics[width=0.95\columnwidth]{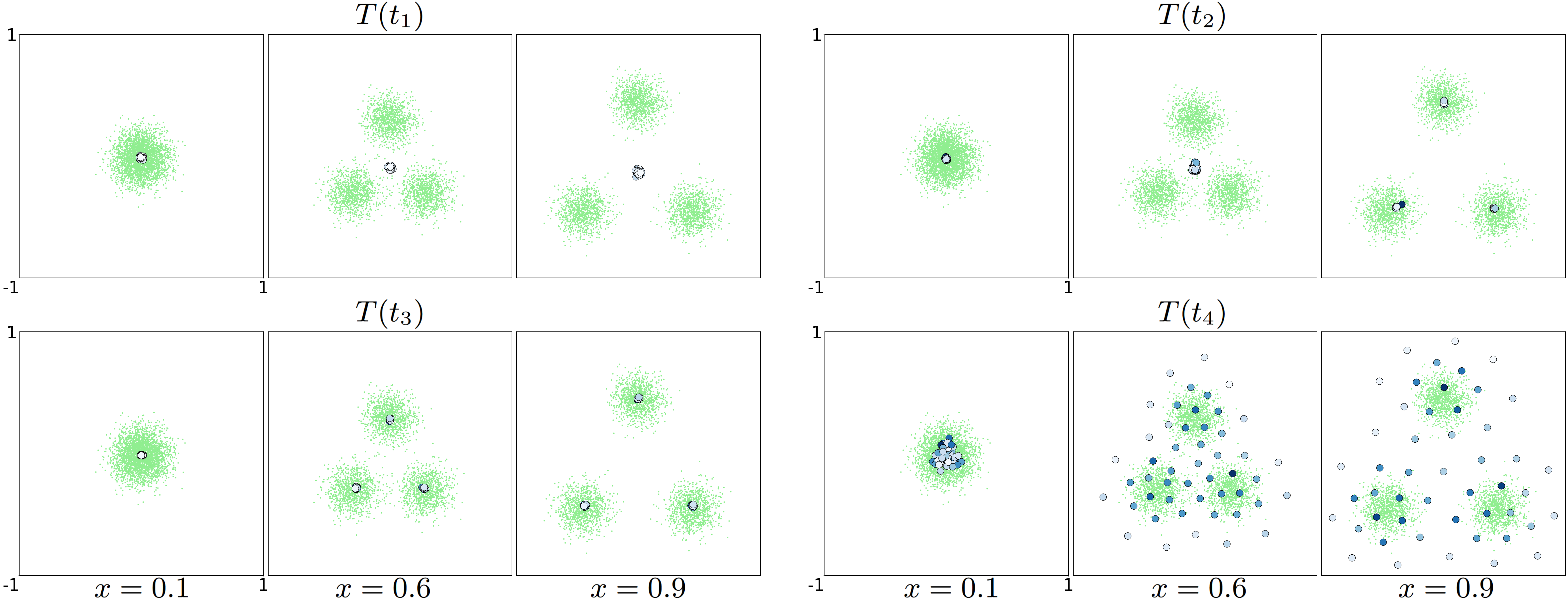}
    \caption{\textbf{Conditional phase transitions with $t_{1} < t_{2} < t_{3} < t_{4}$.} Results for a conditional version of the dataset in Figure \ref{fig:amcl_comparison}, where the Gaussian moves linearly and increases with the input $x \in [0,1]$. aMCL was trained during $1000$ epochs, using a linear scheduler with $T_{0} = 1.0$, and using $49$ hypotheses. Each subplot group corresponds to the predictions of the model at a given temperature, evaluated at different $x$ values. At temperature $T(t_1)$, the hypotheses are at the barycenter. As temperature decreases they undergo a first phase transition (at temperature $T(t_2)$ for $x=0.9$ and $T(t_3)$ for $x=0.6$), moving toward each Gaussian's barycenter, followed by a second phase transition at $T(t_4)$. 
    We see that earlier splits in the cooling schedule correspond to conditional distributions with higher variances.
    }
    \label{fig:bifurcation}
\end{figure}

First, observe that when the temperature is high, the Boltzmann distribution $q_T$ becomes uniform. Therefore, the soft Voronoi cells merge into a single cell $\cY$, the hypotheses $f_k$ converge toward the barycenter of $\cY$, and they all fuse into a single hypothesis: 
\begin{equation}
    f_k^{\star}(x) = \mathbb{E}_{(X,Y) \sim p(x,y)}[Y \mid X=x]\;, \;\;\;\forall k\in\bn\;.
    \label{eq:barycenter_prop}
\end{equation}
Remarkably, this phenomenon occurs even at finite temperatures (see Appendix, Proposition \ref{propapp:temperature_properties}). 
As the temperature decreases, a phenomenon of \textit{bifurcation} occurs \cite{rose1990statistical, rose1994mapping}. During this process, the hypotheses iteratively split into sub-groups, as shown in Figure \ref{fig:bifurcation}. The virtual number of hypotheses \cite{rose1998deterministic} for each $x$ at a given distortion level is captured by the \textit{conditional rate-distortion function}
\begin{equation}
R_{x}(D^{\star}) \triangleq \min _{\substack{q\in\Delta_n, f \in \mathcal{F}\\ D_x(q,f) \leq D^{\star}}} I_x(\hat{Y} ; Y)\;.
\label{eq:rate_distortion}
\end{equation}
In Eq. \eqref{eq:rate_distortion}, $Y\sim p(y\mid x)$ follows the target distribution, the hypothesis position $\hat{Y}\sim q(f_k\mid x)$ follows a distribution over $\mathcal{Y}$ with $q(f_k \mid x) = \int_{\cY} q(f_k \mid x,y) p(y \mid x) \dx$, $I_x(\hat{Y} ; Y)$ is their mutual information, and $D_x(q,f)=\int_{\mathcal{Y}}\sum_{k} \ell(f_k(x), y) q(f_k \mid x, y) p(y \mid x) \dy$ is the distortion for input $x$.

The rate-distortion function $R_x(D^{\star})$ has the following key properties.

\begin{prop}[Rate-distortion properties]\label{prop:rate_distortion} 
See Proposition \ref{propapp:rate_distortion} in Appendix.

For each $x\in\cX$, let $D_{x}^{\mathrm{max}}$ \eqref{eqapp:dmax} denote the optimal conditional distortion when using a single hypothesis, we have the following results.

\begin{itemize}
\item[(i)] For each $T > 0$, minimizing the free energy 
\begin{equation}
    \mF=D(q_{T}, f) - T H(q_{T})\;,
\end{equation}
over all hypotheses positions $f \in \mathcal{F}(\cX,\cY^n)$ comes down to solving the optimization problem that defines \eqref{eq:rate_distortion} for each $x \in \cX$.
\item[(ii)] $R_x$ is a non-increasing, convex and continuous function of $D^{\star}$, $R_x(D^{\star}) = 0$ for $D^{\star} \geq D_{x}^{\mathrm{max}}$, and for each $x$ the slope can be interpreted as $R^{\prime}_x(D^{\star})=-\frac{1}{T}$ when it is differentiable.
\item[(iii)] For each $x$, $R_x(D^{\star})$ is bounded below by the Shannon Lower Bound (SLB)
\begin{equation}
R_x(D^\star) \geq \mathrm{SLB}(D^\star) \triangleq H(Y) - H(D^\star)\;,
\end{equation}
where $Y \sim p(y \mid x)$ and $H(D^\star)$ is the entropy of a Gaussian with variance $D^\star$.
\end{itemize}
\end{prop}

Part $(i)$ establishes that the rate-distortion function is tightly linked with our problem. Provided that the hypotheses $f_k$ perfectly optimize the soft distortion \eqref{eq:soft_distortion} at any temperature level,  the hypothesis configuration $f$ will follow the optimal parametric curve $(D^{\star}, R_x(D^{\star}))$ for each $x \in \cX$. Part $(ii)$ describes the shape of the parametric curve $(D^{\star}, R_x(D^{\star}))$. The Shannon Lower Bound described in part $(iii)$ is a lower bound on the virtual number of hypotheses reached by aMCL.

\begin{figure}
    \centering
    \begin{minipage}{0.47\textwidth}
        \vspace{-38pt}
        \centering
\includegraphics[width=\linewidth]{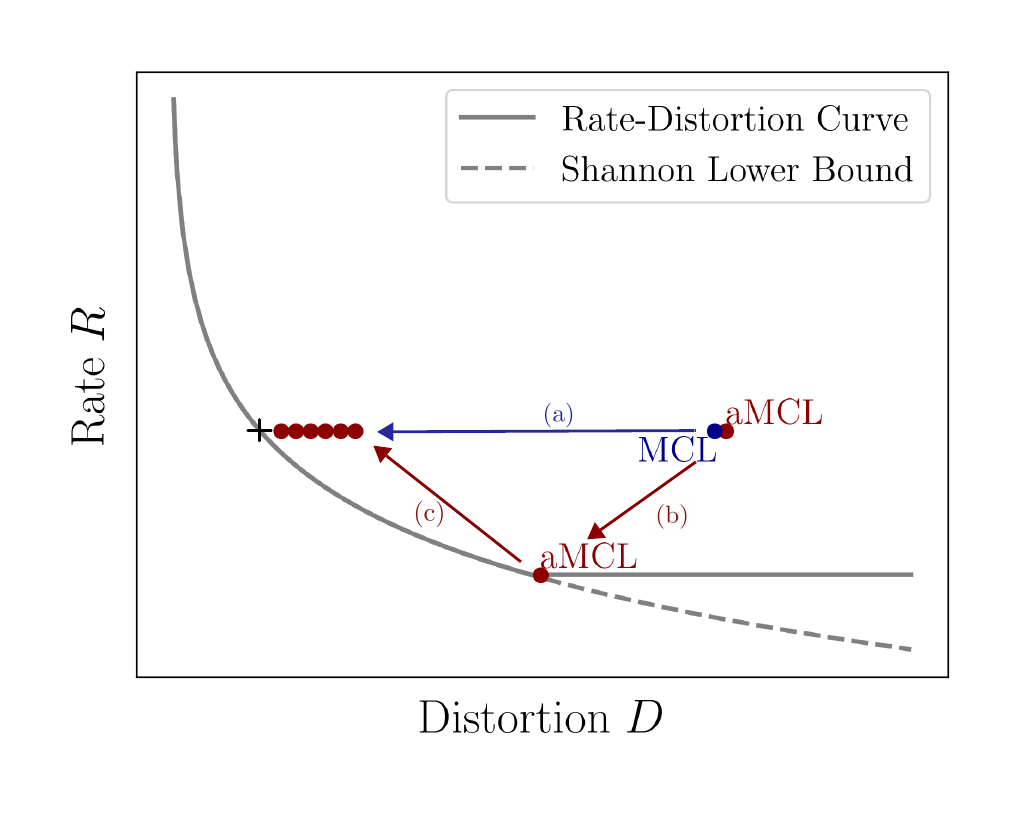}
\caption{\textbf{Illustration of the training trajectory in the Rate-Distortion curve.} Training trajectories of MCL (blue) and aMCL (red) in the case of a single Gaussian. The optimal reachable distortion (`+') is the distortion $D^\star$ satisfying $R(D^\star) = \log_{2}(n)$ (See Section \ref{sec:rate_distortion}).}
    \label{fig:rate_distortion_curve}
    \end{minipage}\hfill
    \begin{minipage}{0.50\textwidth}
        \centering
        \includegraphics[width=\columnwidth]{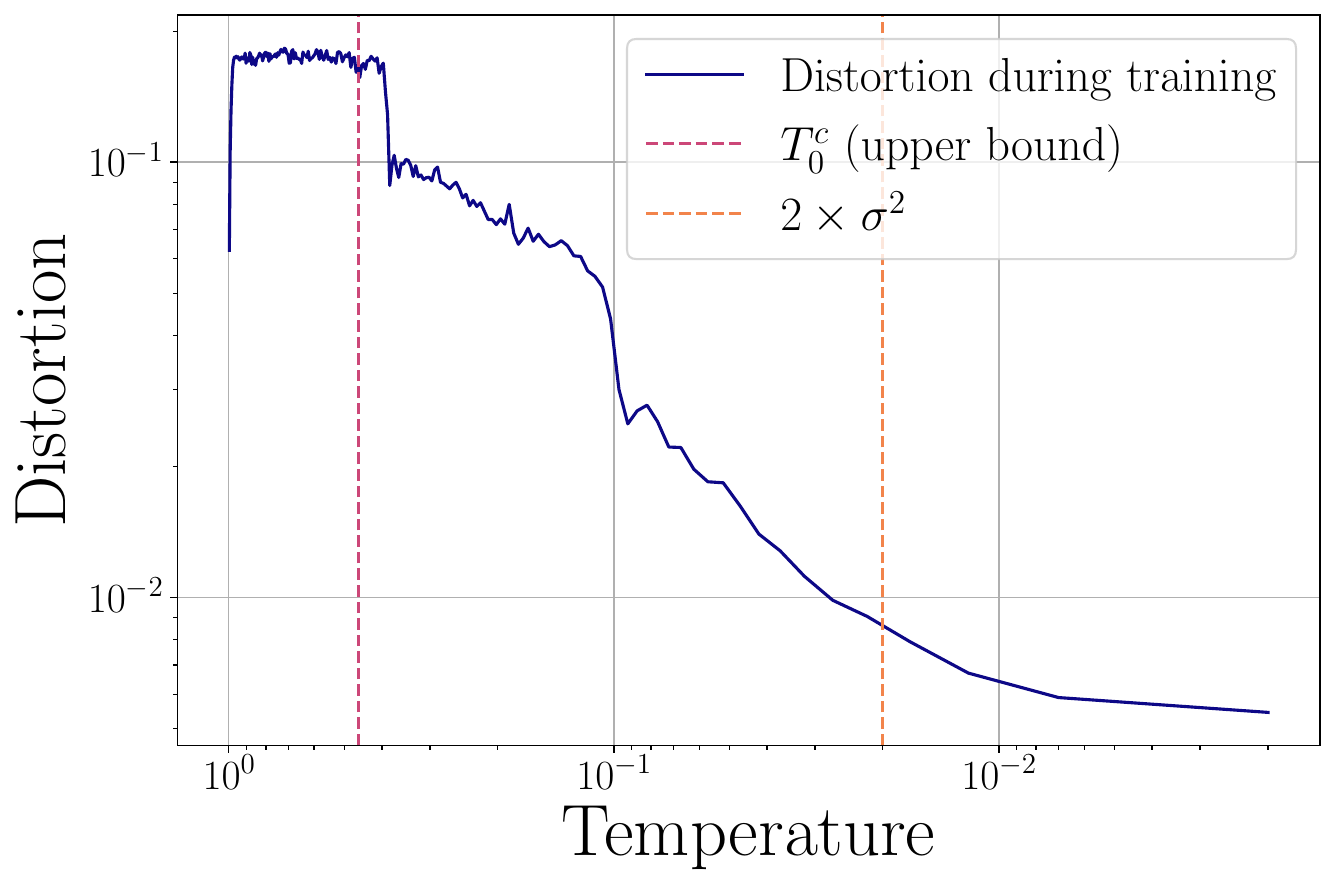} 
    \caption{\textbf{Regimes in the distortion \eqref{eq:distortion} vs. temperature training curve on the setup of Figure \ref{fig:bifurcation}}. At first, the hypotheses converge to the conditional mean. It is followed by a plateau phase where performance stagnates. Transition begins at $T_{0}^c$: the hypotheses migrate toward the barycenter of each Gaussian. Then, they split and we observe a last phase transition. For reference, $2 \sigma^2$ is the critical temperature for a Gaussian with variance $\sigma^2$.
    }
    \label{fig:dist_vs_temp}
    \end{minipage}
\end{figure}

The rate-distortion curve effectively describes the training trajectory of our algorithm, with the deterministic annealing procedure consisting of ascending along this curve \cite{rose1990statistical, rose1994mapping}.
Interestingly, there is a set of critical temperatures at which the hypotheses suddenly split, increasing the number of sub-groups they form. By analogy with statistical physics, the behavior at these points has been coined \textit{phase transitions}\nobreak\,\cite{graepel1997phase}. 
An illustration of the trajectory of MCL and aMCL in the rate-distortion space is shown in Figure \ref{fig:rate_distortion_curve}. We see that MCL evolves along a constant rate $R = \log_{2}(n)$ (in bits) following $(a)$. In contrast, aMCL initially reaches the critical state at $D^{\star} = D_x^{\mathrm{max}}$ following $(b)$. After the transition, the trajectory of aMCL returns to the maximal rate following $(c)$. We expect the optimization at a lower rate to be simpler and this training trajectory to provide a better initialization for the set of hypotheses compared to the vanilla MCL.

\subsection{Phase transitions}
\label{sec:phase_transition}

During training, as the temperature decreases, the hypotheses $f$ undergo a sequence of phase transitions at specific critical temperatures. The right tool to exhibit these critical temperatures is the Hessian of the free energy $f \mapsto \mathscr{F}(f,q_{T})$. Transitions occur when the minimum of $\mathscr{F}$ is no longer stable, and it can be shown \cite{rose1990statistical} that this relates to the eigenvalues of the following block-diagonal covariance matrix:
\begin{equation}
C_{k,k}(f, q | x) = \int_\cY ( f_k(x)-y ) (f_k(x)-y)^t \; q(y \mid f_k, x) p(y \mid x) \dy\;,
\label{eq:covariance}
\end{equation}

where $q(y \mid f_k, x)$ denotes the posterior probability of assigning a point $y$ to the hypothesis $k$, calculated using Bayes’s rule \cite{rose1998deterministic}. At high temperatures, all hypotheses merge into the conditional barycenter of the distribution \eqref{eq:barycenter_prop}. In this setting, all the matrices $C_{k,k}$ are equal to the data covariance matrix $C(x)\triangleq \mathbb{C}\mathrm{ov}_{(X,Y) \sim p(x,y)}[Y \mid X=x]$, $\mathscr{F}$ has a unique minimizer, and the stability of this global optimum is conditioned on the strict positivity of the Hessian of $\mathscr{F}$. The first critical temperature $T_0^c$ is defined as the first temperature for which the Hessian of $\mathscr{F}$ is no longer positive definite.

This temperature is connected to $D_x^{\max}$, the vanishing point of the rate-distortion function $R_x(D^\star)$, which also corresponds to the optimal 1-hypothesis distortion \cite{berger2003rate}. Indeed, $R_x(D^\star)$ measures the virtual number of hypotheses, so that the first splitting of the hypotheses from a single point to several groups will coincide with the moment when $R_x(D^\star)>0$. We summarize these observations in the following Proposition \ref{prop:conditional_critical_temp}, which is illustrated in Figure \ref{fig:dist_vs_temp}.

\begin{prop}[First critical temperature]
\label{prop:conditional_critical_temp}
See Definition \ref{def:first_critical_temperature}, Propositions \ref{propapp:additional_distortion_prop} $(ii)$ and \ref{prop:firstcritical_temp} in Appendix. Let $\lambda_{\max}(\cdot)$ denote the maximum eigenvalue of a matrix, $D_x^{*}(T)\triangleq\mathrm{inf}_{f\in\cF} D_x(q_{T}, f)$, $D^{*}(T)\triangleq\mathrm{inf}_{f\in\cF} D(q_{T}, f)$, and $D_{\mathrm{max}}\triangleq\int_{x \in \cX} D_x^{\mathrm{max}}p(x)\dx$. Then the two following properties hold.
\begin{itemize}
    \item[(i)] $D_x^*$ and $D^*$ are non-decreasing functions of $T$ and admit generalized inverses (with the convention $g^{-1}(\beta)=\mathrm{inf} \{ \alpha \mid g(\alpha) \geq \beta \}$ for the generalized inverse of a function $g$).
    \item[(ii)] The conditional (resp. non-conditional) first critical temperature $T_0^c(x)$ (resp. $T_0^c$) satisfy
    \begin{align}
        T_0^c(x) &\triangleq (D_x^{*})^{-1}(D_x^{\mathrm{max}})=2\lambda_{\mathrm{max}}(C(x))\;, \label{eq:first_equation}
        \\
        T_0^c &\triangleq (D^{*})^{-1}\left( D_{\mathrm{max}}\right) \leq 2 \sup_{x \in\cX} \lambda_{\mathrm{max}}(C(x))\;.
        \label{eq:second_equation}
    \end{align}
\end{itemize}
\end{prop}

After the first transition, an interesting phenomenon occurs for some distributions $p(x,y)$. Instead of splitting in all directions, the hypotheses $f_k$ continue to form a small number of subgroups. The hypotheses $f_k$ may undergo many phase transitions before they all split apart from each other: this is illustrated in Figure \ref{fig:bifurcation}. Generally, this recursive splitting reaches an end: under mild conditions \cite{koch2016shannon}, there is a critical temperature below which the hypotheses all separate from each other. This defines the last critical temperature $T_\infty^c$. Remarkably, $T_\infty^c$ is associated with the Shannon Lower Bound: the temperature at which $R_x(D^\star)$ hits the lower bound $\mathrm{SLB}(D^{\star})$ corresponds to the moment when the hypotheses $f_k$ completely separate from each other (see Theorem 1 and 3 in \cite{rose1994mapping}). 

\section{Experimental validation}\label{sec:experiments}

In this Section, we experimentally investigate the advantage of our algorithm in practical settings. Specifically, we evaluate it on the standard UCI benchmark in Section \ref{sec:uci}, and we apply it to the challenging task of speech separation in Section \ref{sec:ss}.

\subsection{UCI datasets}\label{sec:uci}
\subsubsection{Setup}
\label{sec:setup}
\textbf{General setup.} We followed the experimental protocol described by \cite{hernandez2015probabilistic} for the UCI benchmark \cite{dua2017uci}. Specifically, we used the official train-test splits, with 20 folds except for the Protein dataset, which is split into 5 folds, and the Year dataset, which uses a single fold. 

\textbf{Baselines.} In our result tables, we also include data from three baseline methods detailed in Table 1 of Lakshminarayanan \textit{et al.}'s paper \cite{lakshminarayanan2017simple} (`Deep Ensembles'), which serves as a reference. These baselines include Probabilistic Back Propagation \cite{hernandez2015probabilistic} (denoted `PBP'), and Monte Carlo Dropout \cite{gal2016dropout} (denoted `MC-dropout'). As additional baselines, we include the standard score-based MCL (\textit{e.g.}, \cite{letzelter2023resilient}). We also include the Relaxed-MCL variant \cite{rupprecht2017learning} with $\varepsilon = 0.1$. The impact of $\varepsilon$ is discussed in Appendix \ref{sec:impact_epsilon}. Our method (aMCL), was trained with an exponential scheduler of the form $T(t) = T_{0} \rho^{t}$, with $\rho = 0.95$ and $T_{0} = 0.5$. Comparison with a linear scheduler is also provided in Appendix \ref{sec:impact_scheduler}. Both aMCL and Relaxed-MCL were trained for 1,000 epochs. Each MCL system was trained with $n = 5$ hypotheses.

\textbf{Metrics.} 
We computed the following metrics on the UCI datasets. Let $d$ denote the squared Euclidean distance: $d(\hat{y}_i, y_i) = \lVert y_i - \hat{y_i} \rVert^{2}$, and $N$ the number of samples in each dataset. 
The RMSE $(\downarrow)$ is defined as $\mathrm{RMSE} = \sqrt{\frac{1}{N} \sum_{i} d(\hat{y}_i, y_i)}$, where $\hat{y}_i$ denotes the estimated conditional mean. The latter was computed with $\sum_{k=1}^{n} \gamma_k(x_i) f_k(x_i)$ for the MCL variants. The results of this experiment are summed up in Table \ref{tab:uci_distortion} (Distortion), and also in Table \ref{tab:uci_rmse} (RMSE). Rows are ordered by dataset size $N$, with the intensity of the grey color proportional to $N$ (excluding the Year dataset).

\subsubsection{Results}

\begin{table}[h]
\caption{\textbf{Results on UCI regression benchmark datasets comparing Distortion.} Experimental setup is described in Section \ref{sec:setup}. Relaxed-WTA results were computed with $\varepsilon = 0.1$ which strikes a good tradeoff between RMSE and Distortion (see Table \ref{table:epsilon_study} in Appendix). The rows are ordered by dataset size $N$. Best results are in \textbf{bold}, second bests are \underline{underlined.}}
    \begin{center}
    \resizebox{0.8\columnwidth}{!}{
    \begin{tabular}{l ccc|l}
    \toprule
    & \multicolumn{3}{c}{Distortion ($\downarrow$)} & \\
    Datasets & Relaxed-WTA ($\varepsilon = 0.1$) & MCL & aMCL & $N$ \\
    \midrule
\rowcolor{black!35} Year & 9.09 $\pm$ NA & \underline{4.82 $\pm$ NA} & \textbf{4.46 $\pm$ NA} & 515345 \\
\rowcolor{black!30} Protein & 1.67 $\pm$ 0.16 & \underline{0.80 $\pm$ 0.02} & \textbf{0.77 $\pm$ 0.03} & 45730 \\
\rowcolor{black!7.82} Naval & \textbf{4.21e-7 $\pm$ 2.36e-7} & 1.84e-6 $\pm$ 2.42e-6 & \underline{5.37e-7 $\pm$ 3.83e-7} & 11934 \\
\rowcolor{black!6.27} Power & 2.95 $\pm$ 0.91 & \underline{2.31 $\pm$ 0.49} & \textbf{2.18 $\pm$ 0.64} & 9568 \\
\rowcolor{black!5.37} Kin8nm & \underline{9.32e-4 $\pm$ 7.97e-5} & 1.00e-3 $\pm$ 1.47e-4 & \textbf{6.81e-4 $\pm$ 8.14e-5} & 8192 \\
\rowcolor{black!1.04} Wine & 0.06 $\pm$ 0.02 & \textbf{0.02 $\pm$ 0.01} & \underline{0.03 $\pm$ 0.01} & 1599 \\
\rowcolor{black!0.7} Concrete & 6.91 $\pm$ 2.81 & \textbf{5.13 $\pm$ 1.23} & \underline{5.71 $\pm$ 1.72} & 1030 \\
\rowcolor{black!0.5} Energy & \underline{0.30 $\pm$ 0.12} & 1.25 $\pm$ 1.25 & \textbf{0.28 $\pm$ 0.09} & 768 \\
\rowcolor{black!0.3} Boston & 3.32 $\pm$ 2.84 & \textbf{2.14 $\pm$ 0.49} & \underline{2.69 $\pm$ 1.39} & 506 \\
\rowcolor{black!0.2} Yacht & \underline{1.34 $\pm$ 0.93} & 3.09 $\pm$ 2.41 & \textbf{1.15 $\pm$ 0.97} & 308 \\
\bottomrule
    \end{tabular}
    }
    \end{center}
    \label{tab:uci_distortion}
\end{table}

\begin{table}[h!]
\caption{\textbf{Results on UCI regression benchmark datasets comparing RMSE.} Best results are in \textbf{bold}, second bests are \underline{underlined.} $^{*}$ corresponds to reported results from \cite{lakshminarayanan2017simple}.}
\begin{center}
\resizebox{\columnwidth}{!}{
\begin{tabular}{l  ccc|cccc|c}
\toprule
& \multicolumn{6}{c}{RMSE ($\downarrow$)}\\
\cmidrule(l{2pt}r{2pt}){2-7}
Datasets & PBP$^{*}$ & MC Dropout$^{*}$ & Deep Ensembles$^{*}$ & Relaxed-WTA ($\varepsilon = 0.1$) & MCL & aMCL & $N$\\
\midrule
\rowcolor{black!35} Year      & \underline{8.88 $\pm$ NA} & \textbf{8.85 $\pm$ NA} & 8.89 $\pm$ NA & 8.97 $\pm$ NA & 9.09 $\pm$ NA & 9.08 $\pm$ NA & 515345 \\
\rowcolor{black!30} Protein   & 4.73 $\pm$ 0.01 & \underline{4.36 $\pm$ 0.04} & 4.71 $\pm$ 0.06 & 4.38 $\pm$ 0.02 & 4.39 $\pm$ 0.10 & \textbf{4.25 $\pm$ 0.02} & 45730 \\
\rowcolor{black!7.82} Naval     & 0.01 $\pm$ 0.00 & 0.01 $\pm$ 0.00 & \textbf{0.00 $\pm$ 0.00} & \underline{1.80e-3 $\pm$ 5.66e-4} & 2.08e-3 $\pm$ 1.18e-3 & \textbf{8.00e-4 $\pm$ 4.04e-4} & 11934\\
\rowcolor{black!6.27} Power     & 4.12 $\pm$ 0.03 & \textbf{4.02 $\pm$ 0.18} & 4.11 $\pm$ 0.17 & \textbf{4.02 $\pm$ 0.18} & 4.18 $\pm$ 0.16 & \underline{4.08 $\pm$ 0.20} & 9568 \\
\rowcolor{black!5.37} Kin8nm    & 0.10 $\pm$ 0.00 & 0.10 $\pm$ 0.00 & \underline{0.09 $\pm$ 0.00} & \textbf{0.08 $\pm$ 0.00} & 0.10 $\pm$ 0.01 & \textbf{0.08 $\pm$ 0.00} & 8192 \\
\rowcolor{black!1.04} Wine      & 0.64 $\pm$ 0.01 & \textbf{0.62 $\pm$ 0.04} & 0.64 $\pm$ 0.04 & \underline{0.63 $\pm$ 0.04} & \underline{0.63 $\pm$ 0.04} & \underline{0.63 $\pm$ 0.04} & 1599 \\
\rowcolor{black!0.7} Concrete  & 5.67 $\pm$ 0.09 & \textbf{5.23 $\pm$ 0.53} & 6.03 $\pm$ 0.58 & \underline{5.28 $\pm$ 0.58} & 6.02 $\pm$ 0.65 & 5.47 $\pm$ 0.67 & 1030 \\
\rowcolor{black!0.5} Energy    & 1.80 $\pm$ 0.05 & 1.66 $\pm$ 0.19 & 2.09 $\pm$ 0.29 & \underline{1.64 $\pm$ 0.36} & 2.53 $\pm$ 0.99 & \textbf{1.35 $\pm$ 0.97} & 768 \\
\rowcolor{black!0.3} Boston    & 3.01 $\pm$ 0.18 & \underline{2.97 $\pm$ 0.85} & 3.28 $\pm$ 1.00 & \textbf{2.85 $\pm$ 0.72} & 3.54 $\pm$ 1.16 & 3.05 $\pm$ 0.91 & 506 \\
\rowcolor{black!0.2} Yacht     & \textbf{1.02 $\pm$ 0.05} & \underline{1.11 $\pm$ 0.38} & 1.58 $\pm$ 0.48 & 2.52 $\pm$ 1.04 & 3.28 $\pm$ 1.39 & 1.62 $\pm$ 0.53 & 308 \\
\bottomrule
\end{tabular}
}
\end{center}
\label{tab:uci_rmse}
\end{table}

\newpage

\textbf{Comparison of aMCL and MCL.} We can observe that aMCL performs comparably or outperforms vanilla MCL in most settings, especially for large dataset sizes, both in terms of distortion and RMSE. This outcome supports the claims made in the paper and is especially promising, given that the temperature scheduler was not specifically optimized for each dataset. 

\textbf{Comparison of aMCL and standard UCI baselines.} We observe that aMCL performs on par with, and in some cases exceeds, standard baselines on the RMSE metric. This is noteworthy, as aMCL is not explicitly optimized for RMSE during training—its primary focus is on quantization. While perfect quantization would naturally result in optimal RMSE, achieving low RMSE is not the main objective of aMCL. For example, the RMSE performance of MCL is slightly worse across most datasets in those experiments.

\textbf{Comparison with Relaxed-MCL.} Finally, we compare aMCL and Relaxed-MCL, since aMCL can be interpreted as an adaptative version of Relaxed-MCL. Our results indicate that aMCL generally outperforms Relaxed-MCL in terms of distortion across nearly all datasets. However, we also observe that Relaxed-MCL demonstrates strong performance in terms of RMSE. We attribute these findings to the bias of Relaxed-MCL toward the conditional barycenter of the target distribution, which seems to improve RMSE at the expense of increased distortion. This trade-off arises because RMSE evaluates the accuracy of the barycenter estimation, while distortion measures the quantization performance. The trade-off between distortion and RMSE can be adjusted by tuning the value of $\varepsilon$. Further analysis of this parameter is presented in Appendix \ref{sec:impact_scheduler}, where we show that aMCL strikes a good balance between these two metrics.

These results strongly support the use of aMCL as a quantization algorithm in practical settings. To further evaluate its effectiveness, we also apply aMCL to a more challenging task, namely speech separation.

\subsection{Application to speech separation}\label{sec:ss}

Speech separation consists of isolating each speaker's signal from a mixture in which they are simultaneously active.
This task is of major interest for automatic speech processing applications \cite{marti2012automatic,li2021espnet, von2019all}. In these experiments, we explore the application of MCL and the proposed aMCL to the task of speech separation. An extensive description of the experiments is proposed in Appendix \ref{sec:apx:exp_setting}.

\subsubsection{Experimental setting}

\textbf{General purpose}.
Speech separation consists in obtaining the source signals $y_1,\ldots,y_m\in\mathbb{R}^l$ from a mixture $x=\sum_{s=1}^{m} y_s$. Hence, the task is to provide estimates $\hat{y}_1,\ldots,\hat{y}_m$ of the isolated speech tracks from the mixture $x$.

\textbf{Dataset}.
Source separation experiments are conducted on the Wall Street Journal dataset \cite{hershey2016deep} (WSJ0-mix), a standard benchmark for speech separation. We focus on the 2- and 3- speaker mixture scenarios, with each scenario including 20,000 training, 5,000 validation, and 3,000 testing mixtures.

\textbf{Model architecture}.
The source separation task is solved using the Dual-Path Recurrent Neural Network (DPRNN) architecture \cite{luo2020dual}. DPRNN has been extensively used in speech separation, as it strikes a good balance between performance and number of trainable parameters \cite{wang2023tf,subakan2021attention}.

\textbf{Separation metrics}.
We use the Scale-Invariant Signal-to-Distortion Ratio (SI-SDR) to measure the separation quality \cite{vincent2006performance,le2019sdr}.
There is an ambiguity in finding the best assignment between predicted and active sources. The PIT SI-SDR loss \cite{kolbaek2017multitalker} initially addresses this issue. We propose to use MCL and our new variant aMCL to perform this matching (See Appendix \ref{apx:ssep}). 

\subsubsection{Results}\label{sec:ss_results2}

\begin{wraptable}{r}{0.49\textwidth}
\vspace{-13pt}
\caption{\textbf{2- and 3- speaker source separation} with PIT (topline), MCL, aMCL and Relaxed-WTA ($\varepsilon=0.05$). PIT SI-SDR metric $(\uparrow)$ on the WSJ0-mix eval set. Results over three training seeds, with mean and standard deviation reported.
}
    \centering
    \begin{tabular}{lll}
        \toprule        
            Method &  2 speakers & 3 speakers\\
            \midrule
            PIT & 16.88 \hfill$\pm$ 0.10 & 10.01 \hfill$\pm$ 0.04\\
            \midrule
        MCL & 16.30 \hfill$\pm$ 0.59 & 10.06 \hfill$\pm$ 0.21\\
        Relaxed-WTA & 16.70 \hfill$\pm$ 0.08 & 9.43 $\pm$ 0.21\\
        aMCL & 16.85 \hfill$\pm$ 0.13 & 10.00 \hfill$\pm$ 0.21\\
    \bottomrule
    \end{tabular}
    \label{tab:ssep_perfs}
\end{wraptable}
\textbf{Comparing PIT, MCL, Relaxed-WTA and aMCL.} Table \ref{tab:ssep_perfs} presents the source separation results in the 2- and 3-speaker scenarios. First, aMCL demonstrates performance equivalent to or better than MCL. 
Both methods can be used for the separation task. However, we observed that MCL is subject to hypothesis collapse for some training seeds, while aMCL is more robust to initialization. This translates into a lower inter-seed standard deviation for aMCL. Second, we observe the advantage of aMCL over Relaxed-WTA, which is consistent with our previous analysis of the barycenter bias of this method. Third, aMCL performs equivalently to PIT in both scenarios. Note that by using MCL or aMCL, the number of predictions $n$ could exceed the number of sources $m$. This could be leveraged to improve the separation metrics (cf. Appendix \ref{sec:apx:hyp_num}). Finally, aMCL improves the algorithmic complexity of PIT from $\mathcal{O}(m^3)$ to $\mathcal{O}(mn)$ (cf. Appendix \ref{sec:apx:ssep_perf}). These results make aMCL stand as a good alternative to PIT.

\textbf{Observing phase transition.} When the metric is the Euclidean distance, the theoretical analysis of Section \ref{sec:phase_transition} and the synthetic experiments (see Figure \ref{fig:dist_vs_temp}) have highlighted a phenomenon of phase transition for aMCL.
Here, we analyze the validation loss trajectory as a function of the temperature for different initial temperatures, 
and using an exponential scheduler. The curves with the two higher initial temperatures in Figure \ref{fig:apx:transi} exhibit a plateau until a given temperature. 
After this critical point, the loss decreases.
Although the SI-SDR is non-Euclidean, this behavior resembles that observed for the Euclidean metric. This is detailed in Section \ref{sec:apx:transitions} of the Appendix.

\section{Conclusion}\label{sec:conclusion}

This article introduces aMCL, a novel training method that combines deterministic annealing and the Winner-Takes-all training scheme to address two key issues of MCL: hypothesis collapse and convergence toward a suboptimal local minimum of its quantization objective. We provide a detailed analysis of aMCL's training dynamics. Moreover, drawing on statistical physics and information theory, we provide insights into the trajectory of the aMCL predictions during training. In particular, we exhibit a phase transition phenomenon and establish its connection to the rate-distortion curve. We validate our analysis with experiments on synthetic data, on the UCI datasets, and on a real-world speech separation task. This demonstrates that aMCL is a theoretically grounded alternative to MCL in diverse settings. Future work includes a detailed analysis of the temperature schedule's impact, the derivation of performance bounds, a thorough examination of our algorithm's convergence, particularly at finite temperature,
as well as an evaluation of its generalization capabilities on out-of-distribution samples.

\textbf{Limitations.}  First, aMCL introduces a temperature schedule: this is a challenging hyperparameter tightly linked to the optimizer and its learning rate. The derivation of optimal schedules is left to future work. Second, annealing requires a slow temperature schedule to maintain model performance. This potentially leads to longer training times.

\textbf{Broaden impact.} This paper introduces research aimed at progressing the field of Machine Learning. While our work has numerous potential societal implications, we believe there are no specific consequences that need to be emphasized in this paper.

\section{Acknowledgments}

This work was funded by the French Association for Technological Research (ANRT CIFRE contract 2022-1854) and Hi! PARIS through their PhD in AI funding programs, and was performed using HPC resources from GENCI–IDRIS (Grant 2021-AD011013406R1). We are grateful to the reviewers for their insightful comments.

\bibliography{references}

\newpage
\appendix

\section*{Organisation of the Appendix}

The Appendix is organized as follows. Appendix \ref{sec:apx:theoretical_analysis} presents the theoretical analysis of our algorithm. It begins with the introduction of the notations and the definition of the training scheme in Appendices \ref{sec:apx:notation} and \ref{sec:apx:training_scheme} respectively. This is followed by an interpretation of the algorithm in terms of entropy-constrained alternate optimization in Appendix \ref{sec:apx:alternating_optimization}. Appendix \ref{sec:apx:rate_distortion_curve} provides an analysis of the training dynamics of the algorithm in relation to rate-distortion theory, and Appendix \ref{sec:apx:first_phase_transition} discusses phase transitions. Connection with the literature and additional discussions are provided in Appendix \ref{sec:connection}. Additional details and results from experiments on synthetic data and UCI datasets are provided in Appendices \ref{seccapp:synthetic} and \ref{sec:apx:uci}. Finally, Appendix \ref{apx:ssep} offers an extensive description of the Source Separation experiment, including the impact of the number of hypotheses in Appendix \ref{sec:apx:hyp_num}, and the analysis of phase transitions for this task in Appendix \ref{sec:apx:transitions}.

\section{Theoretical analysis}\label{sec:apx:theoretical_analysis}

\subsection{Notations and motivations}
\label{sec:apx:notation}

Following the main paper notations, let $\mathcal{X}\subset\bR^{d_1}$ and $\mathcal{Y}\subset \bR^{d_2}$ denote the input and target spaces respectively. We will assume that $\cX$ and $\cY$ are bounded. We note $p(x,y)$ the continuous density of a joint data distribution on $\mathcal{X} \times\ \mathcal{Y}$. Let $f = (f_1, \dots, f_n)\in\cF_{\Theta}\subset\cF(\cX,\cY^n)$ denote the hypothesis models, which are families of $n$ neural networks with parameters in $\Theta$. Likewise, let $\gamma=(\gamma_1, \dots, \gamma_n)\in\cF_{\Theta^\prime}\subset\cF(\cX,[0,1]^n)$ denote the score models, with parameters in $\Theta^\prime$. We will sometimes write $\cF=\cF(\cX,\cY^n)$. 

We denote by $\Delta_n$ the set of discrete distributions on $n$ items conditioned by points on $\cX\times\cY$ (typically representing a soft assignment of a target to the hypotheses). We denote by $H(q)$ the entropy of a distribution $q$. If $q\in\Delta_n$, we write $H(q)=-\int_\cXY\sum_{k=1}^n q(f_k |x,y) \log q(f_k |x,y) p(x,y) \dx \dy$. If $Z$ is a random variable with density $p(z)$ and $Y$ is another random variable, we define the differential entropy of $Z$ as $H(Z) = - \mathbb{E}_{Z}[\mathrm{log}(p(Z))]$ and its conditional entropy as $H(Z \mid Y) = -\mathbb{E}_{(Z,Y)}[\mathrm{log}(p(Z \mid Y))]$. Mutual information $I(Z,Y)$ between two random variables $Z$ and $Y$ is defined as $I(Z;Y)=H(Z) - H(Z \mid Y)$.

Let $t\in\bN$ denote a training step, and $t_{\mathrm{epoch}}\in\bN\cup\{\infty\}$ denote the number of training epochs. Let $T: t \mapsto T(t) \geq 0$ be a temperature schedule that decreases with the training step and verifies $\mathrm{lim}_{t \rightarrow t_{\mathrm{epoch}}} T(t) = 0$. Note that we consider the temperature to be constant across an epoch, with linear or exponential decrease between epochs. Let $\ell: \mathcal{Y}^{2} \rightarrow \mathbb{R}_{+}$ denote an underlying loss function. We will restrict our analysis to the Euclidean squared distance $\ell(\hat{y},y) = \lVert y - \hat{y} \rVert^2$ unless otherwise stated. For any vector $z$, we note $z^t$ as its vector transposition. Gradient descent involves a learning rate schedule, that we note $\lambda_t$.

Multiple Choice Learning (MCL) aims at training the hypotheses $(f_{1}, \dots, f_{n})$ such that, for each input $x \in \mathcal{X}$, the hypothesis predictions $(f_{1}(x), \dots, f_{n}(x))$ provide an efficient \textit{quantization} of the conditional distribution $p(y \mid x)$ \cite{rupprecht2017learning, letzelter2023resilient, letzelter24winner}. The quality of the quantization can be evaluated using the \textit{distortion}: 
 \begin{equation}D(f) = \int_{\cX\times\cY} \min_{k\in\bn} \ell(f_k(x),y) p(x,y) \dx\dy\;,
 \label{eqapp:distortion}
 \end{equation} which can be seen as a generalization of the conditional distortion \cite{pages2003optimal}. The two definitions coincide when $p(y \mid x)$ is independent of $x$. Minimizing \eqref{eqapp:distortion} is NP-hard \cite{aloise2009np,dasgupta2008hardness,mahajan2012planar}. The K-means algorithm, which is the standard approach for this task, can take an exponential number of steps before converging \cite{vattani2009k}, and is only guaranteed to find a local minimum in general settings \cite{du2006convergence,bourne2015centroidal}. The same applies to WTA: due to its greedy nature, it can be trapped, at any point of its training trajectory, into a local minimum that it will never manage to escape. One common issue is the \textit{hypothesis collapse} \cite{pages2016pointwise,emelianenko2008nondegeneracy}, a situation in which some of the hypotheses are left unused, therefore leading to a suboptimal configuration (see Figure \ref{fig:amcl_comparison}).

To mitigate these issues, we introduce an annealed version of MCL. It can be seen as an input-dependent version of deterministic annealing \cite{rose1990statistical, rose1992vector}. Our training scheme is described in Appendix \ref{sec:apx:training_scheme}.

 \subsection{Training scheme of aMCL}
\label{sec:apx:training_scheme}
Let $(x, y)$ be a training example sampled from $p$ at step $t$. Our proposed annealed MCL training process is defined through the following subsequent steps:
\begin{enumerate}
    \item Perform a forward pass through $f_1(x), \dots, f_n(x)$. 
    \item Compute the Boltzmann distribution $q_{T(t)}$ defined for each $f_k$ by 
    \begin{equation}\label{eqapp:awta_assignation}
    q_{T(t)}(f_k \mid x, y) \triangleq \frac{1}{Z_{x,y}} \exp \left(-\frac{\ell(f_k(x),y)}{T(t)}\right)\;, \;\; Z_{x,y} \triangleq \sum_{k = 1}^{n} \exp \left(-\dfrac{\ell(f_k(x),y)}{T(t)} \right)\;,
    \end{equation}
    and \textbf{detach} it from the computational graph.
    \item Perform a gradient step on the loss 
    \begin{equation}\label{eqapp:awta_loss}
    \mathcal{L}^{\mathrm{aWTA}} \triangleq \sum_{k=1}^{n} q_{T(t)}(f_k \mid x,y) \ell(f_{k}(x),y)\;,
    \end{equation} 
\end{enumerate}

Note that we used the same scoring loss $\mathcal{L}_{\mathrm{scoring}}$  \eqref{eq:scoring_risk} as in rMCL, the score-based method in \cite{letzelter2023resilient}.

In this paper, we extend the distortion $D(f)$ to account for a distribution $q$ on the hypotheses, and define the resulting \textit{soft distortion} as 
\begin{equation}\label{eqapp:soft_distortion}
    D(q, f) = \int_{\cX\times\cY} \sum_{k=1}^n\ell(f_k(x),y)) q(f_k \mid x, y) p(x,y) \dx\dy\;.
\end{equation}

When $q=q_{T(t)}$, the soft distortion $D(q_{T(t)},f)$ corresponds to the expectation of the aWTA loss \eqref{eqapp:awta_loss}. 

We will now introduce a \textit{decoupling} assumption, which states that the family of models $\cF_{\Theta}$ is expressive enough to encompass the global minimizer of the soft distortion. Noting $D_x(q,f)$ the integrand of \eqref{eqapp:soft_distortion} over the $\cX$ integral, we can formalize this assumption as follows.

\begin{assumption}[Decoupling]
We assume that the model family is perfectly expressive, \textit{i.e.} $\cF_\Theta=\cF$. 
\label{asm_expressivity}
\end{assumption}

Note that under this assumption, the global minimizer of the soft distortion $D(q,f)$, minimizes the integrand of \eqref{eqapp:soft_distortion} for all $x\in\cX$:
\begin{equation}\label{eqapp:distortion_decoupling}
    \forall q\in\Delta_n, \quad \inf_{f\in\cF_{\Theta}} D(q,f) = \inf_{f\in\cF(\cX,\cY^n)} D(q,f) = \int_\cX \inf_{f\in\cF(\cX,\cY^n)} D_x(q,f)\;p(x)\dx\;.
\end{equation}

In the subsequent Sections, our goal is to analyze the aMCL training scheme and justify its design. Specifically, we show that annealing simplifies the optimization problem defined by \eqref{eqapp:distortion}. In Appendix \ref{sec:apx:alternating_optimization}, we first focus on studying the properties of the soft distortion $D(q,f)$. 

In the following, if $x \in \mathcal{X}$ is fixed and when the context is clear, we will omit the dependency on the input and denote the hypotheses position $(f_1, \dots, f_n) \in \mathcal{Y}^n$.

\subsection{Soft assignation and entropy constraint}\label{sec:apx:alternating_optimization}

In this Section, we rewrite the assignation and optimization steps of aMCL as an entropy-constrained block optimization of the soft distortion $D(q,f)$. Noting $H_{T}=H[q_{T}]$, we formalize this result with the following Theorem. 

\begin{theorem}[Entropy constrained distortion minimization] The assignation \eqref{eqapp:awta_assignation} and optimization \eqref{eqapp:awta_loss} steps of aMCL correspond to a block coordinate descent on the soft distortion. For all $k\in\bn$, 
\begin{align}
    & \;\;\;\; q \leftarrow \underset{{\substack{q \in \Delta_n \\ H(q) \geq H_{T(t)}}}}{\mathrm{argmin}} D(q,f)\;,\label{eqapp:distortion_optim}\\
    & \;\;\;\; f_k \leftarrow f_k - \lambda_t\nabla_{f_k} D(q,f)\;.\label{eqapp:gradient_update}
\end{align}
\label{propapp:block_optimization}
\end{theorem}

Theorem \ref{propapp:block_optimization} is a corollary of the following Proposition, which provides additional intuition into the dynamic of aMCL.

\begin{prop}[aMCL training dynamic]\label{propapp:awta_dynamic}
For all $T>0$, $f\in\cF$, strictly positive $q\in\Delta_n$, $x\in\cX$ and $y\in\cY$, the following statements are true. 
\begin{alignat*}{3}
    &(i) \quad  \underset{{\substack{q\in\Delta_n \\ H(q) \geq H_T}}}{\mathrm{argmin}} D(q,f) =q_T\;, \quad && q_T(f_k | x, y) = \frac{\exp\left(-\ell(f_k(x),y\right)/T)}{\sum_{s=1}^n \exp\left(-\ell(f_s(x),y\right)/T)}\;, \quad && \forall k \in \bn \\
    &(ii) \quad  \argmin_{f\in\cF} D(q,f)= f^{\star}\;, \quad && f^{\star}_k(x)=\frac{\int_\cY y \; q(f_k^{\star} | x, y)p(y \mid x)\dy}{\int_\cY q(f_k^{\star} | x, y)p(y \mid x)\dy}\;, \quad && \forall k \in \bn \\
   & (iii)  \quad  \nabla_{f_k} D(q,f) = \gamma^{\star}_k (f_k-f^{\star}_k)\;, \;\;\;\; && \gamma^{\star}_k= \int_\cY q(f_k^{\star} | x, y)p(y \mid x)\dy\;, \quad && \forall k \in \bn
\end{alignat*}
\end{prop}

Part $(i)$ states that the $\mathrm{softmin}$ operator is the solution of the entropy-constrained minimization of the soft distortion. Therefore, gradient-based optimization of $D(q,f)$ along axis $q$ is unnecessary, and we can directly use the closed-form expression in $(i)$ for each temperature level $T(t)$, or equivalenty each corresponding entropy level $H_{T(t)}$. Consequently, the optimization of $D(q,f)$ reduces to the gradient-based minimization $D(q_{T(t)}, f)$ along axis $f$ for each temperature level $T(t)$. 

Part $(ii)$ states that a necessary condition for the minimization of the soft distortion is that each $f_k$ is a soft barycenter of the assignation distribution $q_{T(t)}$ for each temperature $T$. Part $(iii)$ states that each gradient update moves $f_k$ towards this soft barycenter $f^\star_k$, and that the update speed depends on the probability mass $\gamma^\star_k$ of the points softly assigned to $f_k$. Together, these results describe the training dynamics of aMCL. Let us prove these results.

\begin{proof}
$(i)$ First observe that under the decoupling assumption, minimizing $D(q,f)$ amounts to minimizing the integrand $D_x(q,f)=\int_\cY\sum_{k=1}^n \ell(f_k(x),y)) q(f_k \mid x,y)\pxy \mathrm{d}y$ for each $x\in\cX$ (see \eqref{eqapp:distortion_decoupling}). When $x$ is fixed, the distribution $q$ becomes a tuple of $n$ scalars $(q_1,\ldots,q_n)$, and the Lagragian of \eqref{eqapp:distortion_optim} writes $$\cL = D(q,f) - T[H(q)-H_T] + \lambda \left(\sum_{k=1}^n q_k-1\right)\;.$$
Using the first-order necessary optimization conditions, we find that
\begin{align*}
    \ell(f_k,y)+T[\log(q_k)+1]+\lambda&=0\,,\quad \quad
     H_{T}-H[q]=0\;,\\
    \sum_{k=1}^n q_k - 1 &= 0\;.
\end{align*}
From this it immediately follows that $\lambda=1+1/\sum_{k=1}^n\exp\left(-\frac{\ell(f_k,y)}{T}\right)$ is a normalization factor, $q_k=\exp\left(-\frac{\ell(f_k,y)}{T}\right)/\sum_{s=1}^n\exp\left(-\frac{\ell(f_s,y)}{T}\right)$ is the Boltzmann distribution, and $H_T=H[q_{T}]$ correspond to its entropy.\\

$(ii)$ Using the same reasoning than in $(i)$ we set $x\in\cX$. Then each $f_k\in\cY$ becomes a simple vector. Observing that $\nabla_{f_k} D_x(q,f)$ vanishes at the minimum of $D_x(q,f)$, a necessary condition to minimize $D(q,f)$ is
\begin{align}
    \nabla_{f_k} D_x(q,f) &= \int_\cY \frac{\partial \ell}{\partial f_k} \left(f_k,y\right) q(f_k | x,y)\pxy \dy=0\label{eq:distance}\\
    &= 2 \int_\cY (f_k-y)q(f_k |x, y)\pxy \dy=0\;.\\
    \end{align}
Then, as we assumed $q>0$, we have $\int_\cY q(f_k \mid x,y)\pxy \dy > 0$, and
    \begin{equation}
    f_k(x) = \frac{\int_\cY y \; q(f_k \mid x,y)\pxy \dy}{\int_\cY q(f_k \mid x,y)\pxy \dy}\;. 
\end{equation}
$(iii)$ It follows immediately from $(ii)$ that
$$\nabla_{f_k} D_x(q,f) = \left(\int_\cY q(f_k | x,y)\pxy \dy \right) \left(f_k - \frac{\int_\cY y \; q(f_k | x,y)\pxy \dy}{\int_\cY q(f_k | x,y)\pxy \dy}\right) = \gamma_k^{\star}(f_k - f_k^{\star})\;.$$
\end{proof}

Before further analyzing the training dynamics of aMCL, let us pause to examine a few properties of the soft distortion $D(q,f)$. First, observe that the hard distortion $D(f)$ given by \eqref{eqapp:distortion} is a particular case of soft distortion $D(q,f)$ where $q(f_k|x,y)=\bI[k\in \argmin_{s\in\bn} \ell(f_s(x), y)]$. Second, $\inf_{f\in\cF}D_x(q_{T},f)$ and $\inf_{f\in\cF}D(q_{T},f)$ are non-decreasing functions of $T$. Third, entropy-constrained soft distortion minimization \eqref{eqapp:distortion_optim} is equivalent to the minimization of the free energy
\begin{equation}\label{eqapp:free_energy}
    \mF(q,f) = D(q,f) - TH(q)\;,    
\end{equation}
a quantity defined in statistical physics, that will be analyzed in more detail in the following Section. Moreover, optimal free energy has a closed-form formula in our case. We group these observations in the following Proposition.

\begin{prop}[Additional properties of the distortion and the free energy]\label{propapp:additional_distortion_prop}
    For all $T>0$, $f\in\cF$, $q\in\Delta_n$, $x\in\cX$ and $y\in\cY$, the following statements are true. 
    \begin{alignat*}{2}
    &(i) \;\;  &&q(f \mid x, y) = (\; \bI[k\in \argmin_{s\in\bn} \; \ell(f_s(x), y)] \;)_{k\in\bn} \Rightarrow \forall q'\in\Delta_n,\;D(q, f)\leq D(q', f)\;.\\
    &(ii) \;\; &&D_x^*:T\mapsto\inf_{f\in\cF}D_x(q_{T},f) \text{\;and\;} D^*:T\mapsto\inf_{f\in\cF}D(q_{T},f) \text{\; are non-decreasing on $]0,\infty[$\;.}\\
    &(iii) \;\; &&\min_{q\in\Delta_n}\mF(q,f) = \mF(q_T,f)=-T \int_\cXY \log \sum_{k=1}^n\exp\left(-\frac{\ell(f_k(x),y)}{T}\right)p(x,y)\mathrm{d}x\mathrm{d}y\;.
\end{alignat*}
\end{prop}

\begin{proof}
$(i)$ Notice that $\argmin_{s\in\bn} \ell(f_s(x),y)$ selects the hypothesis $f_s$ with lowest loss $\ell(f_s(x),y)$. Therefore, for all $q'\in\Delta_n$, omitting the $x$ dependency in the notations,
$$\sum_{k=1}^n q(f_k | y) \ell(f_k,y) = \min_{k\in\bn}\ell(f_k,y)=\sum_{k=1}^n q'(f_k | y)\min_{s\in\bn} \ell(f_s,y)\leq\sum_{k=1}^n q'(f_k | y) \ell(f_k,y)\;.$$
We conclude by multiplying by $p(x,y)$ and integrating over $\cX\times\cY$. \\
$(ii)$ Let us first show that for any $f \in \mathcal{F}$, $T \mapsto D_x(q_{T},f)$ is non-drecreasing. For that, let us define $\varphi(T) \triangleq \mathbb{E}_{k \sim q_T(f_k \mid x,y)}[\ell(f_k(x),y)] = \sum_{k = 1}^{n} q_T(f_k \mid x,y) \ell(f_k(x),y)$ for each $T>0$, $x,y\in\cX\times\cY$, and $f\in\cF$. 
$\varphi$ is a differentiable function of $T$, with derivative
\begin{align*}
   \varphi^{\prime}(T) &= \frac{1}{T^2} \left[ \sum_{k=1}^{n} q_T(f_k \mid x,y) \ell(f_k(x),y)^{2} - \left( \sum_{s=1}^{n} q_T(f_k \mid x,y) \ell(f_s(x),y) \right)^{2} \right] \\
   &= \frac{1}{T^2} \mathbb{V}_{k \sim q_T(f_k \mid x,y)}[\ell(f_k(x),y)] \geq 0\;.
\end{align*}
Therefore, $\varphi$ is non-decreasing. By the growth of the integral, we deduce that $T \mapsto D_x(q_{T},f)$ is non-decreasing.
\\
Let $T' \geq T>0.$ Then $D_x(q_{T'},f) \geq D_x(q_{T},f)\geq \inf_{f'\in\cF} \;D_x(q_{T},f')$ for all $f\in\cF$, and by definition of the infimum, we have $\inf_{f\in\cF} D_x(q_{T'},f) \geq \inf_{f\in\cF} D_x(q_{T},f)$. 
So $D_x^*$ is a non-decreasing function of $T$, and the same reasoning applies to $D^*$.
\\
$(iii)$ There are two equalities to prove. First observe that $$\mF(q,f)=\underbrace{D(q,f)-T[H(q)-H_T]}_{\textrm{Lagragian of \eqref{eqapp:distortion_optim}}} - \underbrace{T\; H_T}_{\textrm{independent of $q$}}.$$
We deduce
$$\argmin_{q\in\Delta_n} \; \mF(q,f)=\argmin_{q\in\Delta_n} \;(\;D(q,f)-T[H(q)-H_T]\;) = q_T\;,$$
which establishes the first equality. Then we conclude by substituting the definition of $q_T$ into $\mF(q_T,f)$ and simplifying the expression. Again, we drop the $x$ dependency in the notations for readability.
\begin{align*}
    \mF(q_T,f) &= \int_\cXY \sum_{k=1}^n q_T(f_k \mid y)\left[\ell(f_k,y)+T\log q_T(f_k \mid y)\right]p(x,y)\mathrm{d}x\mathrm{d}y \\
    &= \int_\cXY \sum_{k=1}^n \underbrace{q_T(f_k|y)}_{\textrm{sums to 1}}\left[\underbrace{\ell(f_k,y)-T\frac{\ell(f_k,y)}{T}}_{=0}-\underbrace{T\log\sum_{s=1}^n\exp\left(-\frac{\ell(f_s,y)}{T}\right)}_{\textrm{independent of k}}\right]p(x,y)\mathrm{d}x\mathrm{d}y\\
    &= -T \int_\cXY \log \sum_{k=1}^n\exp\left(-\frac{\ell(f_k,y)}{T}\right)p(x,y)\mathrm{d}x\mathrm{d}y\;.
\end{align*}
\end{proof}

Note that the last equality can be written as:
\begin{equation}
    \mF(q_T,f) = -T \int_\cXY \log (Z_{x,y})p(x,y)\dx\dy\; = \int_\cX \mF_x(q_T,f) p(x)\dx\;,
    \label{eqapp:free_energy_partition}
\end{equation}
where we define the conditional free energy $\mF_x(q_T,f)$ as
\begin{equation}
    \mF_x(q_T,f) \triangleq -T \int_\cY \log (Z_{x,y})p(y \mid x)\dy\;.
    \label{eqapp:conditional_free_energy_partition}
\end{equation}

\subsection{Rate distortion curve}
\label{sec:apx:rate_distortion_curve}

We have established in the previous Section that aMCL moves the hypotheses $f_k$ towards the soft barycenter of soft Voronoi cells. We now describe the impact of temperature cooling on the position of these soft barycenters. 

At high temperature, we observe that the soft barycenters (hence the hypotheses) converge towards the conditional barycenter $\bE[Y | X=x]$ and fuze. When temperature decreases, they iteratively split into sub-groups. To capture the virtual number of hypotheses, we introduce the \textit{conditional rate-distortion function}
\begin{equation}
R_x(D^{\star}) \triangleq \min _{\substack{q\in\Delta_n, f\in\cF \\D_x(q,f) \leq D^{\star}}} I_x(\hat{Y} ; Y)\;,
\label{eqapp:conditional_rate_distortion}
\end{equation}
where the target $Y\sim p(y\mid x)$, the hypothesis position $\hat{Y}\sim q(f_k\mid x)$ follows a distribution over $\mathcal{Y}$ with $q(f_k \mid x) = \int_{\cY} q(f_k \mid x,y) p(y \mid x) \dx$, and $I_x(\hat{Y} ; Y)$ is their mutual information. 

The following Proposition shows that \eqref{eqapp:conditional_rate_distortion} corresponds in fact to the same optimization problem as minimizing the conditional free energy \eqref{eqapp:conditional_free_energy_partition}, and therefore as minimizing the entropy constrained distortion \eqref{eqapp:distortion_optim} conditionally on $x$. It then describes the shape of the parametric curve $(D^\star,R_x(D^\star))$, and provides a lower bound of the rate distortion. For this purpose, we introduce for each $x\in\cX$ 
\begin{equation}\label{eqapp:dmax}
D_x^{\mathrm{max}} \triangleq \inf_{f_1\in\cY}\int_\cY \ell(f_1,y) p(y \mid x)\dy = \int_\mathcal{Y} \ell(\mathbb{E}[Y | X=x], y) \, p(y \mid x) \, \dy \;,
\end{equation}
the optimal conditional distortion when using a single hypothesis.

\begin{prop}[Rate-distortion properties]\label{propapp:rate_distortion} 

For each $x\in\cX$, we have the following results.

\begin{itemize}
\item[(i)] For each $T > 0$, minimizing the free energy 
\begin{equation}
    \mF=D(q_{T}, f) - T H(q_{T})\;,
\end{equation}
over all hypotheses positions $f \in \mathcal{F}(\cX,\cY^n)$ comes down to solving the optimization problem that defines \eqref{eq:rate_distortion} for each $x \in \cX$ under decoupling Assumption \ref{asm_expressivity}.
\item[(ii)] $R_x$ is a non-increasing, convex and continuous function of $D^{\star}$, $R_x(D^{\star}) = 0$ for $D^{\star} \geq D_{x}^{\mathrm{max}}$, and for each $x$ the slope can be interpreted as $R^{\prime}_x(D^{\star})=-\frac{1}{T}$ when it is differentiable.
\item[(iii)] For each $x$, $R_x(D^{\star})$ is bounded below by the Shannon Lower Bound (SLB)
\begin{equation}
R_x(D^\star) \geq \mathrm{SLB}(D^\star) \triangleq H(Y) - H(D^\star)\;,
\label{eq:slb}
\end{equation}
where $Y \sim p(y \mid x)$ and $H(D^\star)$ is the entropy of a Gaussian with variance $D^\star$.
\end{itemize}
\label{propapp:free_energy}
\end{prop}

These results are well-known properties of the rate-distortion function \cite{davisson1972rate, berger2003rate, gray1989source, merhav2011rate, blahut1987principles}. 

\begin{proof}
$(i)$ Under the decoupling assumption \ref{asm_expressivity}, optimizing $\mF$ comes down to optimizing $\mF_x$ for each $x \in \cX$. Consequently, we can use the known relationship between the rate-distortion curve and free energy minimization \cite{berger2003rate, gray1989source, rose1994mapping}.

$(ii)$ See \cite{berger2003rate, gray1989source}.

$(iii)$ See \cite{shannon1959coding}.
\end{proof}

A key characteristic of the SLB is its asymptotic tightness in the low distortion regime for a broad range of probability distributions \cite{iacobelli2016asymptotic}. For some distributions, such as the Gaussian case, the SLB and the rate-distortion curve align perfectly for $D \leq D_x^{\mathrm{max}}$. 

The Rate-Distortion interpretation of aMCL applies for each $x$, but defining a notion of global rate that is independent of the input is more challenging. This is because the distortion optimal distortion at a given temperature level $T$, $\inf_{f \in \mathcal{F}} D_x(q_T,f)$, can vary across inputs $x$. This is left for future work.

\subsection{First phase transition}
\label{sec:apx:first_phase_transition}
In the previous Section, we mentioned that aMCL predictions fuze at high temperature into the conditional barycenter $\bE[Y | X=x]$. Moreover, these fuzed predictions iteratively split as the temperature (or equivalently the distortion) decreases during training. With each split, the number of sub-groups formed by the hypotheses increases. The number of sub-groups, measured in bits, is captured by the rate-distortion $R_x(D^\star)$. In this Section, we focus on the critical temperature corresponding to the first of these splits.

The predictions fuze into a single point at high temperatures so that the number of sub-groups is 1 and $R_x(D^\star)=0$ in this regime. The first splitting therefore occurs when $R_x(D^\star)>0$. Correspondingly, we define the first critical temperature in Definition \ref{def:first_critical_temperature}.
\begin{definition}
\label{def:first_critical_temperature}
The first critical temperature $T_0^c(x)$ for each $x$ and the defined global variant $T_0^c$ are defined as:
\begin{align}
  T_0^c(x) &\triangleq \inf \left\{ T \mid \inf_{f \in \mathcal{F}} D_x(q_T, f) \geq D_x^{\max} \right\}\;, \label{eq:first_critical_temp_x}\\
  T_0^c &\triangleq \inf \left\{ T \mid \inf_{f \in \mathcal{F}} D(q_T, f) \geq D_{\max} \right\}\;, \label{eq:first_critical_temp}
\end{align}
in the conditional and the non-conditional setting, where $D_{\mathrm{max}} \triangleq \int_{\cX} p(x) D_x^{\max} \dx$ and $D_x^{\max}$ is defined in \eqref{propapp:rate_distortion}.
\end{definition}
Note that we have also equality in the definition $T_0^c(x) = \inf \left\{ T \mid \inf_{f\in\cF} D_x(q_T, f) = D_x^{\max} \right\}$, since $\inf_{f\in\cF} D_x(q_T, f)\leq D_x^{\max}$. Moreover, recalling that $D_x^*:T \mapsto \mathrm{inf}_{f\in\cF} D_x(q_{T}, f)$ is a non-decreasing function (Proposition \ref{propapp:additional_distortion_prop} $(ii)$), we can equivalently define $T_0^c(x)$ as the generalized inverse $T_0^c(x)=(D_x^*)^{-1}(D_x^{\mathrm{max}})$ (\textit{e.g.}, Definition 1 in \cite{embrechts2013note}). The same observations hold for $T_0^c$.

We also know that aMCL's predictions implicitly minimize the free energy \eqref{eqapp:free_energy} (see Proposition \ref{propapp:free_energy}, part $(i)$). Therefore, the conditional barycenter $\bE[Y | X=x]$ is stable as long as the Hessian of the free energy is positive definite. This observation can alternatively be used to define the first critical temperature.  

Interestingly, all these definitions are equivalent, as shown by the next Proposition. For this purpose, let us introduce the following covariance matrices:
$$C_{k,k}(f, q | x) = \int_\cY ( f_k(x)-y ) (f_k(x)-y)^t \; q(y \mid f_k, x) p(y \mid x)\; \dy\;,$$
where $q(y \mid f_k, x)$ denotes the posterior probability of assigning the point $y$ to the hypothesis $k$, calculated using Bayes’s rule \cite{rose1998deterministic}. At high temperatures, all hypotheses merge into the conditional barycenter of the distribution \eqref{eq:barycenter_prop}, and the matrices $C_{k,k}$ are equal to the data covariance matrix $C(x)\triangleq \mathbb{C}\mathrm{ov}_{(X,Y) \sim p(x,y)}[Y \mid X=x]$.

\begin{prop}[First critical temperature] \label{prop:firstcritical_temp}
    We have the two following results.
    \begin{itemize}
        \item[(i)] We know from \cite{rose1990statistical} that for all $x\in\cX$, $T_0^c(x)=2\lambda_{\max}(C(x))$.
        \item[(ii)] Under decoupling assumption \ref{asm_expressivity}, $T_0^c\leq 2$ $\sup_{x\in\cX} \lambda_{\max}(C(x))$.
    \end{itemize}
\end{prop}

\begin{proof}
    $(i)$ See \cite{rose1990statistical}.
    
    $(ii)$ Let $T=\sup_{x\in\cX} T_0^c(x)$. By definition of $T_0(x)$ and due to Proposition \ref{propapp:additional_distortion_prop} $(ii)$, $\inf_{f\in\cF} D_x(q_T,f) \geq \inf_{f\in\cF} D_x(q_{T_0^c(x)},f) \geq D_x^{\max}$ for all $x\in\cX$. Then, we deduce from \eqref{eqapp:distortion_decoupling}: 
    \begin{align*}
      \inf_{f\in\cF} D(q_T,f) &= \inf_{f\in\cF} \int_\cX D_x(q_T,f)p(x)\dx = \int_\cX \inf_{f\in\cF} D_x(q_T,f)p(x)\dx \geq \int_\cX D_x^{\max}p(x)\dx\;, \\
      \inf_{f\in\cF} D(q_T,f) &\geq D_{\max}\;.  
    \end{align*}
    Using $(i)$, we can write: 
    $$T_0^c = \inf \left\{T \mid \inf_{f\in\cF} D(q_T, f) \geq D_{\mathrm{max}} \right\} \leq T = \sup_{x \in \cX} T_0^c(x)\;.$$
\end{proof}

We now analyze why the predictions converge towards the conditional barycenter $\bE[Y | X=x]$. 
In the next Proposition, we first focus on the asymptotic regime, when the temperature $T$ increases to infinity, and then extend this analysis at finite temperature.

\begin{prop}[Training Dynamics with Temperature]
     For all $x\in\cX$ and all $k\in\bn$, we have the following properties.
    \begin{itemize}
        \item[(i)] If $\;T=\infty\;$ and $f\in\argmin_{f\in\cF}D(q_T,f)$, then $f_k(x)=\bE[Y\mid X=x]$.
        \item[(ii)] The gradient of the soft distortion can be re-written for each $T > 0$ as \begin{equation}\nabla_{f_k} D_x = 2 \int_\cY \sum_{r=0}^{\infty} \frac{1}{r!} \left( -\frac{\|f_k(x) - y\|^2}{T} \right)^r \frac{(f_k(x) - y)}{Z_{x,y}(T)} p(y \mid x)\dy\;.
        \label{eqapp:expansion}\end{equation}
        Therefore, the first order approximation of $\;\nabla_{f_k}D_{x}\;$ in $T$ writes:
        \begin{equation}
        \nabla_{f_k} D_x \underset{T \rightarrow \infty}{=} \frac{2}{n} \int_\cY (f_k(x) - y) p(y \mid x) \dy + o\left(1\right)\;.
        \label{eqapp:asymptotic_expansion}
        \end{equation}
    
    \end{itemize}
    \label{propapp:temperature_properties}
\end{prop}

When $T \rightarrow \infty$, aMCL reduces to standard risk minimization, so that a necessary condition of optimization is that all the hypotheses $f(x)$ are located at the conditional barycenter $\bE[Y | X=x]$ (Proposition \ref{propapp:temperature_properties}, part $(i)$). This conditional barycenter is stable and corresponds to the global minimizer of the aMCL training objective when $T\rightarrow\infty$. We further analyze the \textit{force} that pushes all the hypotheses toward this conditional barycenter, by looking at an expansion \eqref{eqapp:expansion} of $\nabla_{f_k} D$, which provides the direction of the hypotheses updates for each $T > 0$ (Proposition \ref{propapp:temperature_properties}, part $(ii)$). 
As $T \rightarrow \infty$, we see that a global driving force, to which all the data points contribute, is pushing the hypotheses toward the barycenter in \eqref{eqapp:asymptotic_expansion}. As $T$ decreases, local interactions corresponding to the higher-order terms appear and increase in amplitude. They are responsible for the phase transitions. 

\begin{proof}
$(i)$ By definition of the Boltzmann distribution
$$q_T(f_k \mid x,y)=\frac{\exp\left(-\ell(f_k(x),y)/T\right)}{\sum_{s=1}^n \exp\left(-\ell(f_s(x),y)/T\right)} \xrightarrow[T\rightarrow\infty]{} \frac{1}{n}\;.$$
Therefore $D(q_T,f) \xrightarrow[T\rightarrow\infty]{} \frac{1}{n}\sum_{k=1}^n\int_\cY\ell(f_k(x),y)p(y \mid x)\dy$, and all the hypotheses solve the same quantization problem. A necessary condition to minimize $D(q_T,f)$ is then $$\forall k\in\bn,\;f_k(x)=\bE[Y | X=x]\;.$$ 

$(ii)$ We note $Z_{x,y}(T)=\sum_{s=1}^n \exp\left(-\|f_s(x)-y\|^2/T\right)$. Then, accounting for the $\mathrm{stop\_gradient}$ operator in \eqref{eq:awta_loss}, the gradient of the soft distortion writes
$$\nabla_{f_k} D_x (q_T, f) = 2 \int_\cY \frac{\exp\left(-\|f_k(x)-y\|^2/T\right)}{Z_{x,y}(T)} (f_k(x) - y) p(y \mid x)\dy\;.$$
Using the series expansion of the exponential, we have for each $T > 0$,
$$\exp\left( -\frac{\|f_k(x) - y\|^2}{T} \right) = \sum_{r=0}^{\infty} \frac{1}{r!} \left( -\frac{\|f_k(x) - y\|^2}{T} \right)^r\;.$$
We can rewrite the gradient of the soft distortion:
$$\nabla_{f_k} D_x = 2 \int_\cY \sum_{r=0}^{\infty} \frac{1}{r!} \left( -\frac{\|f_k(x) - y\|^2}{T} \right)^r \frac{(f_k(x) - y)}{Z_{x,y}(T)} p(y \mid x)\dy\;.$$
Keeping only the first term, and observing $Z_{x,y}(T)\xrightarrow[T \rightarrow \infty]{}n$ we get:
$$\nabla_{f_k} D_x \underset{T \rightarrow \infty}{=} \frac{2}{n} \int_\cY (f_k(x) - y) p(y \mid x) \dy + o\left(1\right)\;.$$
\end{proof}
This Section focuses on the first phase transition. However, multiple phase transitions may occur during training, as shown in Figure \ref{fig:bifurcation}. It is interesting to note that, for deterministic annealing, the final phase transition occurs when the rate-distortion curve hits the Shannon Lower Bound defined in \eqref{eq:slb} (see \cite{rose1994mapping}).

\section{Connection with the literature and discussion}
\label{sec:connection}
\subsection{Annealing at inference time}

A promising direction for future research consists in using annealing at inference time. With this scheme, aMCL could be used to perform an input-dependent hierarchical clustering \cite{murtagh2012algorithms} at test time, similarly to the idea proposed in \cite{nehme2024hierarchical}. More precisely, we can store the model's parameters at different times of the training schedule (\textit{i.e.}, at several temperature levels). During inference, this allows replaying the temperature cooling for new test samples, by performing forward passes through each of the trained models.

Replaying this trajectory may have several advantages. Indeed, the hypotheses trajectory follows the rate-distortion curve and consequently explores recursively the modes of the distribution as the temperature decays. Crucially, at each critical temperature, when the hypotheses are about to split, they are exactly located at the barycenter of these modes. If we can track these splitting moments, for instance by counting the number of distinct virtual hypotheses at each step of the cooling schedule, we can perform a hierarchical clustering that iteratively uncovers the modes of the distribution.

\subsection{Discussion on Stochastic simulated annealing}

Non-deterministic simulated annealing \cite{hastings1970monte, metropolis1953equation} is a promising research direction due to its strong convergence properties (see Hajek theorem \cite{hajek1988cooling}). It requires to define the state $f$ of the system and the optimization objective $D(f)$. At each step, the state $f$ is updated to a neighbor state $\tilde{f}$ based on a stochastic exploration criterion. The probability of accepting a neighbor state $\tilde{f}$ depends on the objective variation $D(\tilde{f})-D(f)$ and the temperature $T$.

Our objective $D(f)$ corresponds to the Distortion \eqref{eq:distortion}. However, the state of the system can be defined in various ways. In the non-conditional setting, \cite{zeger1992globally} defines the state as the hypothesis positions $(f_k)_{k\in\bn}$ (similarly to the present work), while \cite{hajek1988cooling, hastings1970monte, metropolis1953equation} defines it as the cluster assignation of each dataset sample.

In both cases, we expect that storing and updating this state using neural networks would be costly. Moreover, evaluating $D(\tilde{f})$ requires going through a validation set, which is time-consuming. Further investigation in this direction is left for future research.

\subsection{Comparison with an additional baseline: Relaxed WTA with annealed $\varepsilon$}

Relaxed WTA \cite{rupprecht2017learning} suffers from a bias toward the barycenter. As suggested in \cite{nehme2024hierarchical}, a natural idea to tackle this issue would be to anneal $\varepsilon$ during training. In Figure \ref{fig:amcl_comparison}, we have shown the results of this additional baseline on a Gaussian synthetic dataset, using a linearly decreasing epsilon during training $\varepsilon(t) = \varepsilon_{0} \left(1 - \frac{t}{1000}\right)$ with $\varepsilon_{0} = 0.98$.

\begin{figure}[h!]
    \centering
    \includegraphics[width=0.99\linewidth]{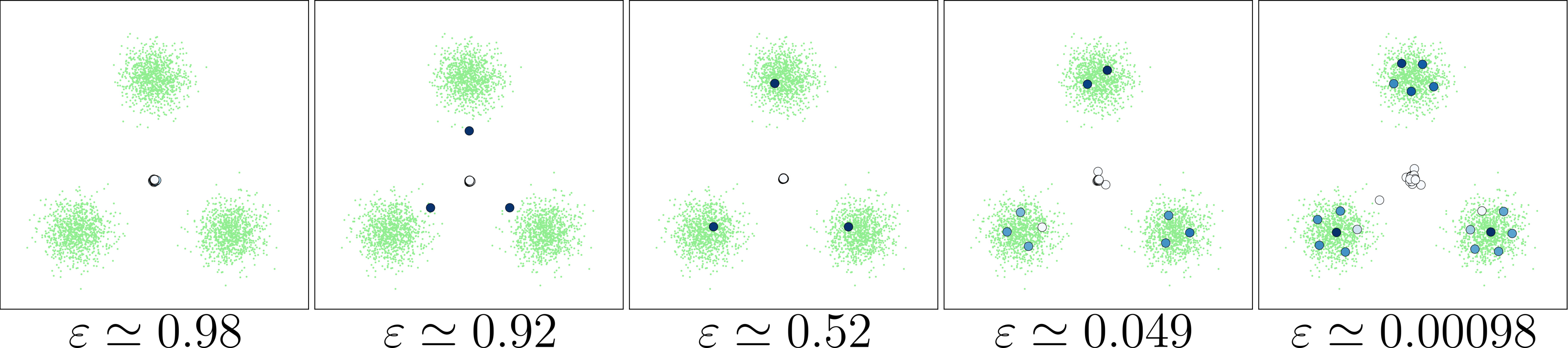}
    \vspace{0.5em}
    \caption{\textbf{Training dynamics of Relaxed-WTA with $\varepsilon$ annealed.} Predictions during training of Relaxed-WTA with $\varepsilon$ annealed linearly over 1000 epochs. We use the same setup as in Figure \ref{fig:amcl_comparison} of the main paper, with 49 hypotheses as shaded blue points (shade proportional to the predicted scores), and the targets as green points. For large $\varepsilon$, the hypotheses converge at the distribution barycenter. As $\varepsilon$ decreases, three \textit{Winners} gradually move towards the barycenter of the Gaussians. When $\varepsilon$ approaches 0, the Winner-Take-All dynamic prevails and some hypotheses escape the barycenter to reach the Gaussian modes. 
    However, the modes are quickly saturated, the unused hypotheses stop receiving gradient, and they remain stuck at the barycenter.
    }
    \label{fig:annealed_epsilon}
\end{figure}

In Figure \ref{fig:annealed_epsilon}, we show its training trajectory when using the same setup as in Figure \ref{fig:amcl_comparison}. All hypotheses initially converge to the barycenter, then the winners gradually move towards the modes as $\varepsilon$ decreases. As $\varepsilon$ approaches $0$, only a few additional hypotheses escape from the barycenter to reach the modes, indicating that annealing does not solve the collapse issue of Relaxed-WTA. 

Results on the UCI datasets confirm this qualitative analysis (Table \ref{table:epsilon_study} and Figure \ref{fig:year_prot}). In Figure \ref{fig:year_prot}, aMCL outperforms the best Relaxed-WTA variants on Distortion.

\section{Experimental details in the synthetic data experiments}
\label{seccapp:synthetic}

This Section provides a detailed description of the synthetic data experiments presented in the paper, particularly the results and visualizations in Figures \ref{fig:amcl_comparison}, \ref{fig:dist_vs_temp} and \ref{fig:bifurcation}. Note that for each of those experiments, a two-hidden-layer Multi-Layer Perceptron (MLP) with $256$ neurons per layer and ReLU activation functions was used, except for the last layer. The final layer of the MLP was duplicated to account for multiple hypotheses $\{f_k(x)\}_{k=1}^{\nhyp}$ (followed by a $\mathrm{tanh}$ activation to restrict the output in the square $[-1,1]^2$) and scores $\{\gamma_k(x)\}_{k=1}^{\nhyp}$ (followed by $\mathrm{sigmoid}$ activation), similar to \cite{letzelter2023resilient, letzelter24winner}, with $n = 49$ hypotheses in this study. At each epoch, 100,000 points were sampled from the corresponding synthetic datasets, and the models were trained for $t_{\mathrm{epoch}} = 1000$ epochs. Note also that the visualization of the output space in Figure \ref{fig:amcl_comparison} and Figure \ref{fig:bifurcation} is restricted to the square $[-1,1]^2$.  

\textbf{Settings of Figure \ref{fig:amcl_comparison}}. In this Figure, the dataset consists of a mixture of three two-dimensional Gaussians located at $\mu_{0} = (-0.5,-0.5)$, $\mu_{1} = (0,0.5)$ and $\mu_{2} = (0.5,-0.5)$, each with a standard deviation of $0.1$.  In this Figure, each model was trained with a Stochastic Gradient Descent optimizer with a constant learning rate of $0.01$. For the Relaxed WTA run, we used $\varepsilon=0.1$, and for the Annealed MCL run, we used an exponential scheduler in the form $T(t) = T_{0} \rho^t$, with $T_0 = 0.6$ and $\rho = 0.99$.

\textbf{Settings of Figure \ref{fig:dist_vs_temp} and Figure \ref{fig:bifurcation}.} In these experiments, the dataset is a conditional variant of that used in Figure \ref{fig:amcl_comparison}, with a ground-truth distribution of the form $p(y \mid x) = \frac{1}{3} \sum_{i=0}^{2} \mathcal{N}(\tilde{\mu}_{i}(x),\sigma^2)$, with $\sigma = 0.1$, $\tilde{\mu}_{i}(x) = x \mu_{i}$ for $x \in [0,1]$, and $\mu_i$ is defined as above for $i \in [\![0,2]\!]$. Here, aMCL was trained with a linear scheduler defined by $T(t) = T_{0}(1 - \frac{t}{t_{\mathrm{epoch}}})$, with $T_{0} = 1.0$. Note that in the results of Figure \ref{fig:bifurcation}, we have $(t_1,t_2,t_3,t_4) = (50,745,870,1000)$ with associated temperature values $(T(t_1),T(t_2),T(t_3),T(t_4)) \simeq (0.950,0.256,0.131,0.001)$. Here, aMCL was trained with Adam optimizer \cite{Kingma-arXiv-2015-Adam}, with a constant learning rate of $0.01$. In Figure \ref{fig:dist_vs_temp}, $T_{0}^{c}$ (upper bound) correspond to the right-hand-side of \eqref{eq:second_equation} and was computed as $2 \lambda_{\mathrm{max}}( \int_{\mathcal{Y}} y y^{t} p(y \mid x = 1) \mathrm{d}y) \simeq 0.46$, approximating the covariance matrix over 1,000 samples. Note that the distortion plotted in the curve of Figure \ref{fig:dist_vs_temp} corresponds to the hard distortion \eqref{eq:distortion} averaged over a validation set of 25,000 samples every $5$ epoch. 

\section{Additional results from UCI Datasets experiments}
\label{sec:apx:uci}

The UCI Regression Datasets \cite{dua2017uci} serve as a standard benchmark for conditional distribution estimation. This Section provides additional details and results from the experiments presented in Section \ref{sec:uci}.

\textbf{Experimental setup.} The experiments were conducted using the protocol from \cite{hernandez2015probabilistic}. Each dataset is divided into 20 train-test folds, except the protein dataset, which had 5 folds, and the Year Prediction MSD dataset which used a single test-set split. We used the same neural network backbone for each baseline: a one-hidden layer MLP with ReLU activation function, containing 50 hidden units except for the Protein and Year datasets, for which 100 hidden units were used. In the WTA models, the final layer of the MLP was duplicated to account for multiple hypotheses $\{f_k(x)\}_{k=1}^{\nhyp}$ (followed by no activation) and scores $\{\gamma_k(x)\}_{k=1}^{\nhyp}$ (followed by sigmoid activation). The WTA-based methods (Relaxed-MCL, MCL, and aMCL) were trained with $\nhyp = 5$. All models were trained with the Adam optimizer over 1,000 epochs with a constant learning rate of $0.01$. The data loading pipeline was adapted from the implementation of \cite{han2022card}. The best models for each dataset are highlighted in bold in Table \ref{tab:uci_rmse} and Table \ref{tab:uci_distortion}, with the second-best models underlined.

\textbf{Baselines.} Table \ref{tab:uci_rmse} of the main paper includes results from three baselines reported from Table 1 of \cite{lakshminarayanan2017simple}, which we use as references for those benchmarks. `PBP' stands for Probabilistic Back Propagation \cite{hernandez2015probabilistic}, and `MC-dropout' corresponds to Monte Carlo Dropout \cite{gal2016dropout}.
Relaxed-MCL was trained with $\varepsilon = 0.1$ in Tables \ref{tab:uci_distortion} and \ref{tab:uci_rmse}. Other values of $\varepsilon$ are experimented in Appendix \ref{sec:impact_epsilon}.

\textbf{Metrics.} We used RMSE and distortion as metrics. RMSE is defined as $\mathrm{RMSE} = \sqrt{\frac{1}{N} \sum_{i=1}^N \| \hat{y}_i - y_i \|^2}$, where $\hat{y}_i$ denotes the estimated conditional mean, defined as $\sum_{k=1}^{\nhyp} \gamma_k(x) f_k(x)$ for the WTA variants, and $N$ is the number of samples in each test set. The distortion of the multi-hypotheses models was computed as $\mathrm{Distortion} = \frac{1}{N} \sum_{i=1}^N \underset{k \in [\![1,\nhyp]\!]}{\mathrm{min}} \| \hat{y}_k -  y_i \|^2$.

\textbf{Temperature scheduler.} For training aMCL in those experiments, an exponential temperature scheduler was used, defined as $T(t) = T_{0} \rho^{t}$, with $\rho = 0.95$ and $T_{0} = 0.5$, where $t$ denotes the current epoch and $t_{\mathrm{epoch}} = 1,000$ is the maximum number of epochs. When the temperature of the exponential scheduler was lower than $5e-4$, we set the training back to the vanilla Winner-takes-all mode.  

\textbf{Evaluation details.} During the evaluation, following \cite{hernandez2015probabilistic}, both input and output variables were normalized using the training data's means and standard deviations. For evaluation, the original scale of the output predictions was restored using the transformation $f_{\theta}^k(x) \mapsto \mu_{\mathrm{train}} + \sigma_{\mathrm{train}} f_{\theta}^k(x)$ where $\mu_{\mathrm{train}}$ and $\sigma_{\mathrm{train}}$ are the empirical mean and standard deviation of the output variable, computed across the training set. 

The results concerning the distortion metric on the UCI datasets are presented in Table \ref{tab:uci_distortion}, and their analysis is provided in Section \ref{sec:uci}.

\subsection{Impact of the scheduler type in aMCL}
\label{sec:impact_scheduler}
The impact of the scheduler type on the performance on the UCI dataset is provided in Table \ref{tab:impact_scheduler_type}. We see that for large dataset sizes, the exponential scheduler outperforms the linear one. Further study on the impact of the speed of the scheduler on the performance is left for further work.

\begin{table}[h!]
\caption{\textbf{Impact of the scheduler type in aMCL.} We display the results on the UCI datasets comparing two types of temperature schedules in aMCL. Here, both use an initial temperature $T_{0} = 0.5$. aMCL (exp) uses an exponential scheduler of the form $T(t) = T_{0} \rho^{t}$, with $\rho = 0.95$, and aMCL (lin) uses a linear scheduler $T(t) = T_0 (1 - \frac{t}{t_{\mathrm{epoch}}})$. The rows are ordered by dataset size $N$.}
    \begin{center}
    \resizebox{1.0\columnwidth}{!}{
    \begin{tabular}{lll||ll|l}
{} & \multicolumn{2}{c}{RMSE ($\downarrow$)} & \multicolumn{2}{c}{Distortion ($\downarrow$)} & \\
\cmidrule(lr){2-3} \cmidrule(lr){4-5}
Datasets & aMCL (exp) & aMCL (lin) & aMCL (exp) & aMCL (lin) & $N$\\
\midrule
\rowcolor{black!35} Year & \textbf{9.08 $\pm$ NA} & 9.20 $\pm$ NA & \textbf{4.46 $\pm$ NA} & 4.50 $\pm$ NA & 515345 \\
\rowcolor{black!30} Protein & \textbf{4.25 $\pm$ 0.02} & 4.26 $\pm$ 0.04 & \textbf{0.77 $\pm$ 0.03} & 0.92 $\pm$ 0.06 & 45730 \\
\rowcolor{black!7.82} Naval & \textbf{8.00e-4 $\pm$ 4.04e-4} & 1.72e-3 $\pm$ 1.71e-3 & \textbf{5.37e-7 $\pm$ 3.83e-7} & 3.97e-6 $\pm$ 1.36e-5 & 11934 \\
\rowcolor{black!6.27} Power & 4.08 $\pm$ 0.20 & \textbf{4.07 $\pm$ 0.16} & \textbf{2.18 $\pm$ 0.64} & 13.24 $\pm$ 2.88 & 9568 \\
\rowcolor{black!5.37} Kin8nm & \textbf{0.08 $\pm$ 0.00} & \textbf{0.08 $\pm$ 0.00} & \textbf{6.81e-4 $\pm$ 8.14e-5} & 2.94e-3 $\pm$ 1.80e-3 & 8192 \\
\rowcolor{black!1.04} Wine & \textbf{0.63 $\pm$ 0.04} & 0.67 $\pm$ 0.05 & \textbf{0.03 $\pm$ 0.01} & 0.09 $\pm$ 0.02 & 1599 \\
\rowcolor{black!0.7} Concrete & 5.47 $\pm$ 0.67 & \textbf{4.99 $\pm$ 0.63} & \textbf{5.71 $\pm$ 1.72} & 10.18 $\pm$ 4.55 & 1030 \\
\rowcolor{black!0.5} Energy & 1.35 $\pm$ 0.97 & \textbf{0.80 $\pm$ 0.33} & \textbf{0.28 $\pm$ 0.09} & 0.55 $\pm$ 0.51 & 768 \\
\rowcolor{black!0.3} Boston & \textbf{3.05 $\pm$ 0.91} & 3.12 $\pm$ 0.68 & \textbf{2.69 $\pm$ 1.39} & 6.18 $\pm$ 2.99 & 506 \\
\rowcolor{black!0.2} Yacht & 1.62 $\pm$ 0.53 & \textbf{0.85 $\pm$ 0.25} & 1.15 $\pm$ 0.97 & \textbf{0.51 $\pm$ 0.37} & 308 \\
\bottomrule
\end{tabular}
}
    \end{center}
    \label{tab:impact_scheduler_type}
\end{table}

\subsection{Impact of $\varepsilon$ in Relaxed WTA}
\label{sec:impact_epsilon}
\begin{figure}[h]
    \centering
\includegraphics[width=\linewidth]{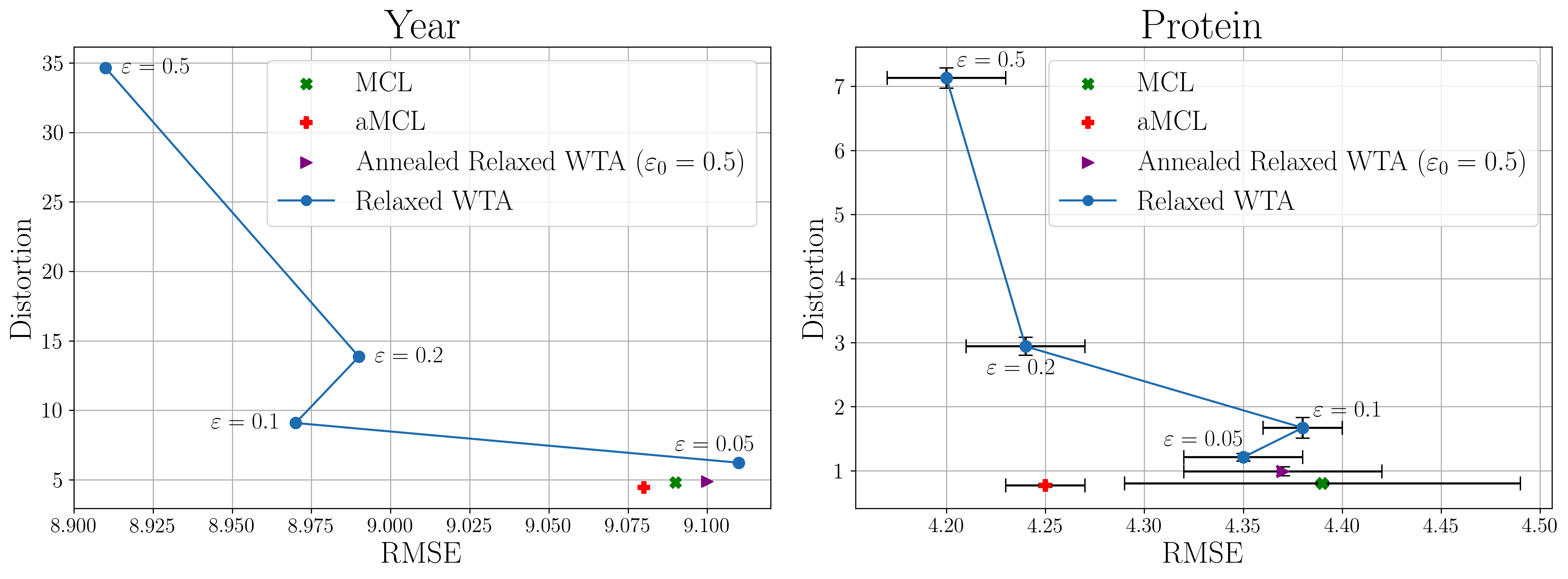}
        \caption{\textbf{(RMSE,Distortion) performance on Year and Protein datasets}, for Relaxed-WTA with $\varepsilon \in \{0.05, 0.1, 0.2, 0.5\}$ (blue points, with black error bars for std), Relaxed-WTA with annealed $\varepsilon$ (purple) and aMCL (red). Blue lines link the Relaxed-WTA score points.  We see that aMCL is below and left of these lines, which indicates a good tradeoff on those two datasets.}
        \label{fig:year_prot}
\end{figure}

RMSE compares the barycenter of the predicted distribution with the target positions. For this metric, Relaxed-MCL outperforms other approaches on the UCI datasets when $\varepsilon$ is high (see Table \ref{table:epsilon_study} and Figure \ref{fig:year_prot}). This outcome is expected, as Relaxed-WTA is biased towards the distribution barycenter under this regime (see Figure \ref{fig:amcl_comparison} of the paper).

However, using RMSE for comparison discards valuable spatial distribution information of the hypotheses. Distortion, defined in \eqref{eq:distortion}, addresses this limitation by measuring quantization performance, which our theoretical analysis is based on. When focusing on the Distortion metric, Table \ref{table:epsilon_study} shows that lower values of epsilon (\textit{e.g.}, $\varepsilon = 0.1$ or annealed epsilon) consistently perform better than for $\varepsilon = 0.5$ across nearly all settings. This improvement is attributed to the reduced bias toward the barycenter in this configuration. Additionally, for Distortion, aMCL (trained with an exponential scheduler) generally outperforms Relaxed-WTA on the UCI datasets for the tested Relaxed WTA variants (see Tables \ref{tab:uci_distortion}, \ref{table:epsilon_study} and Figure \ref{fig:year_prot}). This is especially true on large datasets (Year, Protein).

Overall, aMCL demonstrates a strong balance between Distortion and RMSE compared to the baselines, especially on the largest datasets. Figure \ref{fig:year_prot} further supports this trend, providing additional analysis on Year and Protein, where aMCL, MCL, and Relaxed-MCL are compared across different $\varepsilon$ values in the (RMSE, Distortion) space.

\begin{table}[h!]
\caption{\textbf{Impact of $\varepsilon$ in Relaxed-WTA on UCI regression benchmark datasets comparing RMSE and Distortion.} Best results are in \textbf{bold}, second bests are \underline{underlined}. The annealed version of Relaxed-WTA (R-WTA) uses a linear scheduler starting at $\varepsilon_{0} = 0.5$ and decreasing to 0.
}
    \begin{center}
    \resizebox{1.0\columnwidth}{!}{
    \begin{tabular}{llll||lll|l}
{} & \multicolumn{3}{c}{ RMSE ($\downarrow$)} & \multicolumn{3}{c}{Distortion ($\downarrow$)} & \\
\cmidrule(lr){2-4} \cmidrule(lr){5-7}
{} & R-WTA ($\varepsilon = 0.5$) & R-WTA ($\varepsilon = 0.1$) & R-WTA ($\varepsilon$ annealed) & R-WTA ($\varepsilon = 0.5$) & R-WTA ($\varepsilon = 0.1$) & R-WTA ($\varepsilon$ annealed) & $N$\\
\midrule
\rowcolor{black!35} Year & \textbf{8.91 $\pm$ NA} & \underline{8.97 $\pm$ NA} & 9.10 $\pm$ NA & 34.65 $\pm$ NA & \underline{9.09 $\pm$ NA} & \textbf{4.89 $\pm$ NA} & 515345 \\
\rowcolor{black!30} Protein & \textbf{4.20 $\pm$ 0.03} & 4.38 $\pm$ 0.02 & \underline{4.37 $\pm$ 0.05} & 7.13 $\pm$ 0.16 & \underline{1.67 $\pm$ 0.16} & \textbf{0.99 $\pm$ 0.07} & 45730 \\
\rowcolor{black!7.82} Naval & \textbf{1.47e-3 $\pm$ 7.47e-4} & \underline{1.80e-3 $\pm$ 5.66e-4} & 2.08e-3 $\pm$ 8.49e-4 & 1.13e-6 $\pm$ 1.45e-6 & \textbf{4.21e-7 $\pm$ 2.36e-7} & \underline{7.70e-7 $\pm$ 7.85e-7} & 11934 \\
\rowcolor{black!6.27} Power & \textbf{3.99 $\pm$ 0.16} & \underline{4.02 $\pm$ 0.18} & 4.10 $\pm$ 0.14 & 7.69 $\pm$ 1.26 & \textbf{2.95 $\pm$ 0.91} & \underline{4.28 $\pm$ 2.28} & 9568 \\
\rowcolor{black!5.37} Kin8nm & \textbf{7.68e-2 $\pm$ 1.89e-3} & \underline{7.82e-2 $\pm$ 1.91e-3} & 8.64e-2 $\pm$ 5.25e-3 & 2.82e-3 $\pm$ 2.09e-4 & \underline{9.32e-4 $\pm$ 7.97e-5} & \textbf{9.06e-4 $\pm$ 3.89e-4} & 8192 \\
\rowcolor{black!1.04} Wine & \textbf{0.63 $\pm$ 0.04} & \textbf{0.63 $\pm$ 0.04} & 0.70 $\pm$ 0.06 & 0.19 $\pm$ 0.03 & \textbf{0.06 $\pm$ 0.02} & \underline{0.14 $\pm$ 0.05} & 1599 \\
\rowcolor{black!0.7} Concrete & \textbf{4.91 $\pm$ 0.65} & 5.28 $\pm$ 0.58 & \underline{5.14 $\pm$ 0.56} & 15.08 $\pm$ 5.41 & \textbf{6.91 $\pm$ 2.81} & \underline{7.40 $\pm$ 3.12} & 1030 \\
\rowcolor{black!0.5} Energy & \textbf{1.03 $\pm$ 0.21} & 1.64 $\pm$ 0.36 & \underline{1.48 $\pm$ 0.59} & 0.40 $\pm$ 0.13 & \textbf{0.30 $\pm$ 0.12} & \underline{0.33 $\pm$ 0.10} & 768 \\
\rowcolor{black!0.3} Boston & \underline{2.85 $\pm$ 0.57} & \underline{2.85 $\pm$ 0.72} & \textbf{2.80 $\pm$ 0.69} & 5.95 $\pm$ 2.94 & \textbf{3.32 $\pm$ 2.84} & \underline{3.73 $\pm$ 2.38} & 506 \\
\rowcolor{black!0.2} Yacht & \textbf{1.17 $\pm$ 0.32} & 2.52 $\pm$ 1.04 & \underline{1.59 $\pm$ 0.34} & \textbf{0.63 $\pm$ 0.28} & 1.34 $\pm$ 0.93 & \underline{0.73 $\pm$ 0.49} & 308 \\
\bottomrule
\end{tabular}
}
    \end{center}
\label{table:epsilon_study}
\end{table}

\section{Application to speech separation}
\label{apx:ssep}

This Appendix provides a detailed description of the speech source separation experiments. Speech separation consists of isolating the audio signals of each individual speaker from a mixture in which they are simultaneously active.
This task is of major interest for speech processing applications such as Automatic Speech Recognition \cite{marti2012automatic,li2021espnet}, Speaker Diarization \cite{von2019all}, or singing voice extraction from music \cite{huang2012singing}.
Most of the recent speech separation systems are deep neural networks trained in a supervised setting, where the ground truth of the separated speaker tracks is known \cite{wang2018supervised}. 

\subsection{Task description}
\label{sec:ss_description}

Formally, let $y_1,\ldots,y_m\in\mathbb{R}^l$ denote the raw speech signals from $m$ individual speakers, with $l$ denoting the number of time frames in the corresponding audio recordings. 
Under anechoic (no reverberation) and clean (no background noise) conditions, the mixture can be expressed as $x=\sum_{s=1}^{m} y_s$. Hence, the task is to provide estimates $\hat{y}_1,\ldots,\hat{y}_m$ of the isolated speech tracks from the mixture signal $x$ using a neural network $f_\theta$. The quality of the estimation $\ell(\hat{y}_k, y_s)$ is commonly measured using the audio-domain distance Signal-to-Distortion Ratio (SDR) and its scale-invariant (SI) variant SI-SDR \cite{vincent2006performance,le2019sdr}.
Since there is a fundamental ambiguity concerning the best association $\hat{y}_k\mapsto y_s$, separation systems are trained using the Permutation Invariant Training (PIT) loss:
\begin{equation}\label{eq:apx:pit_loss}
    \cL^{\mathrm{PIT}}_{\mathrm{sep}}(\hat{y},y) =  \min_{\sigma\in\Sigma}\left(\frac{1}{m}\sum_{s=1}^m \ell(\hat{y}_{\sigma(s)}, y_s)\right)\;, 
\end{equation}
where $\Sigma$ is the set of permutations of $\bm$. 

 Instead, we propose to use the MCL framework to solve this assignment problem. 
 Naming $\cL^{\mathrm{MCL}}_{\mathrm{sep}}$ the resulting loss, $\cL^{\mathrm{aMCL}}_{\mathrm{sep}}$ its annealed variant, and $\cL^{\mathrm{Relaxed-WTA}}_{\mathrm{sep}}$ its relaxed variant, we define
\begin{align}
    \cL^{\mathrm{MCL}}_{\mathrm{sep}}(\hat{y},y) &= \frac{1}{m} \sum_{s=1}^m \min_{k\in\bn}\ell(\hat{y}_k, y_s)\;, \label{eq:apx:s_wta_loss} \\
    \cL^{\mathrm{aMCL}}_{\mathrm{sep}}(\hat{y},y)
    &= \frac{1}{m} \sum_{s=1}^m\sum_{k=1}^n q_{T(t)}(\hat{y}_k \mid y_s)\ell(\hat{y}_k, y_s)\;, \label{eq:apx:s_awta_loss} \\
    \cL^{\mathrm{Relaxed-WTA}}_{\mathrm{sep}}(\hat{y},y)
    &= \frac{1}{m} \sum_{s=1}^m\sum_{k=1}^n q_\varepsilon(\hat{y}_k \mid y_s)\ell(\hat{y}_k, y_s)\;, \label{eq:apx:s_ewta_loss}
\end{align}
where $q_\varepsilon(\hat{y}_k \mid y_s)=\begin{cases}
    1-\varepsilon,& \text{if } k=\argmin_{k'\in\bn}\ell(\hat{y}_{k'}, y_s)\\
    \varepsilon/(n-1),              & \text{otherwise}
\end{cases}$.

This gives the separation task a new interpretation: the targets $y_s$ are samples drawn from a common distribution $\pxy$ conditional on the mix $x$, the estimated speech signals $\hat{y}_k=f_k(x)$ are the output of neural networks $f_k$, and the mix $x$ is their common input.

Notice that \eqref{eq:apx:s_wta_loss} is subject to collapse. 
Furthermore, when $n=m$, \eqref{eq:apx:pit_loss} always finds the best assignation, and acts as a topline, while MCL acts as a baseline. 
The goal of these experiments is therefore threefold:
\begin{itemize}
    \item applying our proposed aMCL algorithm to a more realistic and data-intensive setting;
    \item extending the analysis of aMCL to a more general setting where the underlying metric (SI-SDR) is non-Euclidean;
    \item ensuring that MCL is a valid substitute for PIT in the context of source separation, thus opening up a new avenue for MCL research in this domain.
\end{itemize}

\subsection{Experimental settings}
\label{sec:apx:exp_setting}

\subsubsection{Separation model architecture}

We use the Dual-Path Recurrent Neural Network (DPRNN) \cite{luo2020dual} separation architecture for our experiments.
This model follows the encoder-masker-decoder structure, originally proposed in ConvTasNet \cite{luo2019conv}.
Each component of the network is described below:

\textbf{Encoder.}
The encoder transforms the raw audio signal $x\in\mathbb{R}^{l}$, with $l$ the number of time frames, into a sequence of features.
It is implemented as a 1-D convolutional layer with $F$ channels, kernel size $K$, and stride $S$. It outputs the sequence $\mathbf{X}\in\mathbb{R}^{F\times T}$ where $T$ is the number of encoded frames.

\textbf{Masker.}
The masker of the DPRNN model first applies a sliding window of width $W$ to the sequence of encoded features $\mathbf{X}$.
The input sequence is transformed from $\mathbb{R}^{F\times T}$ to $\mathbf{X}^{\prime}\in\mathbb{R}^{F\times P\times W}$, where $P$ is the number of windows.
DPRNN is composed of two recurrent layers, processing the sequence $\mathbf{X}^{\prime}$ over two different directions (dual-path): 
\begin{itemize}
    \item \textit{intra}-chunk: processes the dimension $W$,
    \item \textit{inter}-chunk: processes the dimension $P$.
\end{itemize}

Thus, the model learns local and global dependencies.
The RNN is implemented as a bidirectional Long Short-Term Memory layer (LSTM) with hidden dimension $H$.
The sequence $\mathbf{X}^{\prime}$ is processed by $B$ successive DPRNN blocks.
The output of the last block is denoted $\mathbf{X}^{\prime\prime}\in\mathbb{R}^{H\times P\times W}$.
Before decoding, the sequence is mapped back to the original dimensions using overlap-and-add.
A final convolutional layer maps the sequence $\mathbf{X}^{\prime\prime}$ to the masks $\mathbf{M}\in\mathbb{R}^{n\times F\times T}$, where $n$ is the number of predictions.
The $k$-th source latent sequence $\hat{\mathbf{Y}}_k=\mathbf{M}_k \odot \mathbf{X}$ is inferred by the element-wise product $\odot$ between the $k$-th mask $\mathbf{M}_k$ and the encoded sequence $\mathbf{X}$.

\textbf{Decoder.}
The decoder is implemented as a 1-D transposed convolutional layer with $n$ groups, $F$ channels, and a kernel of size $K$ and stride $S$.
It maps the latent sequences $\hat{\mathbf{Y}}\in\mathbb{R}^{n\times F \times T}$ to the audio domain, and we note $\hat{y}\in \bR^{n\times l}$ the corresponding vector. Therefore, using the MCL framework notations, the output of the prediction model $k$ is $f_k(x)=\hat{y}_k$. 

\textbf{Scoring.} 
We do not use scoring models in these experiments. Indeed, we will focus on the case where $n=m$ and all predictions should be matched to a corresponding source. This ensures a fair comparison with the PIT baseline, which uses the same setting. 

The model configuration used for our experiments is presented in Table \ref{tab:apdx:dprnn_conf}. 
It corresponds to the model configuration from the original paper. The variables $l$, $T$, and $P$, which depend on the unknown input length $l$, are not specified.

\begin{table}[b]
    \centering
    \caption{\textbf{Configuration of the DPRNN model used for source separation experiments.}}
    \begin{tabular}{lcc}
         \toprule
         \textbf{Parameter} & \textbf{Symbol} & \textbf{Value} \\
         \midrule
         Feature dimension & $F$ & 64\\
         Encoder/decoder kernel size & $K$ & 16 \\
         Encoder/decoder stride & $S$ & 8\\
         DPRNN Chunk size & $W$ & 100\\
         Hidden dimension & $H$ & 128\\
         Number of DPRNN blocks & $B$ & 6\\
         \bottomrule
    \end{tabular}
    \label{tab:apdx:dprnn_conf}
\end{table}

\subsubsection{Training configuration}

\textbf{Dataset.}
The source separation experiments are conducted on the Wall Street Journal mix dataset (WSJ0-mix) \cite{hershey2016deep} which is the current benchmark dataset for speech source separation.
It consists of clean utterances of read speech linearly combined to build synthetic speech mixtures.
Several versions of the dataset are available to separate from 2-speaker to 5-speaker mixtures.
In the present work, we focus on the 2- and 3-speaker scenarios.
Each version features 20000, 5000, and 3000 mixtures for training, validation, and testing respectively.
The audio signals are samples at 8kHz.

\textbf{Training objectives.}
Four types of systems are compared in the following experiments:
\begin{itemize}
    \item PIT: training with $\cL^{\mathrm{PIT}}_{\mathrm{sep}}$ defined in \eqref{eq:apx:pit_loss},
    \item MCL: training with $\cL^{\mathrm{MCL}}_{\mathrm{sep}}$ defined in \eqref{eq:apx:s_wta_loss},
    \item aMCL: training with $\cL^{\mathrm{aMCL}}_{\mathrm{sep}}$ defined in \eqref{eq:apx:s_awta_loss},
    \item Relaxed-WTA: training with $\cL^{\mathrm{Relaxed-WTA}}_{\mathrm{sep}}$ defined in \eqref{eq:apx:s_ewta_loss}.
\end{itemize}

We use the SI-SDR score as the separation quality measure
\begin{equation}\label{eq:apx:sisdr}
    \ell(y,\hat{y})=10\log_{\mathrm{10}}\frac{\langle y, \hat{y} \rangle^2}{\|y\|^2\| \hat{y}\|^2 - \langle y, \hat{y} \rangle^2}\;,
\end{equation}
where $\langle\cdot,\cdot\rangle$ denotes the scalar product on $\bR^l$. 

\textbf{Training parameters.} 
The separation models are trained on 5-second audio segments.
Since the utterances from the training set might have a longer duration, 5-second segments are randomly cropped inside the training utterances.
The utterances shorter than 5s are removed.
The batch size is set to 22 to reduce training time compared to the standard setup \cite{luo2019conv,luo2020dual,tzinis2020sudo}, in which a batch size of 4 is used. We verified that our models match the originally reported scores \cite{luo2020dual} when using this original setting.
To analyze robustness to initialization, we train each method with 3 different seeds, and we report the inter-seed average scores and standard deviations. More precisely, the prediction scores are first averaged across the evaluation dataset for each single seed. Then these dataset-wise scores are used to compute the final average and standard deviations.
Each model is trained on $t_{\mathrm{epoch}}=200$ epochs, without early stopping. All the training set is processed during each epoch. Unless otherwise stated, the temperature scheduler is chunk linear: 
\begin{equation}\label{eq:apx:linear_scheduler}
    T(t) = T_0\left(1 - \frac{t}{t_\mathrm{max}}\right) \bI[t < t_{\mathrm{max}}]\;,
\end{equation} 
with initial temperature $T_0\approx0.1$ and $t_\mathrm{max}=100$. Note that at the end of the training, when $t\in[t_\mathrm{max}, t_\mathrm{epoch}]$, annealing is disabled by setting $T(t)=0$ in order to fine-tune the hypotheses on the target MCL objective.

The neural network weights are updated using the Adam optimizer, with a learning rate set to $10^{-3}$.
The learning rate is halved after every 5 epochs with no improvement in the validation metric.
Separation models are trained on Nvidia A40 GPU cards.
The DPRNN has 3.7 million parameters. 

\textbf{Evaluation.}
The separation scores are computed on the evaluation subset of the WSJ0-mix datasets.
Two types of scores are presented:
\begin{itemize}
    \item PIT SI-SDR: the assignment is performed with PIT \eqref{eq:apx:pit_loss} as in standard source separation approaches;
    \item MCL SI-SDR: the assignment is performed with MCL \eqref{eq:apx:s_wta_loss}, and this score can be seen as a form of quantization error.
\end{itemize}

\subsection{Experimental results}

This Section presents the results of the experiments conducted on speech separation with the aMCL framework, and is organized as follows:
\begin{itemize}
    \item Appendix \ref{sec:apx:ssep_perf}: comparison of the separation performance of each approach;
    \item Appendix \ref{sec:apx:hyp_num}: impact of the number of hypotheses on the separation performance with MCL and aMCL;
    \item Appendix \ref{sec:apx:transitions}: analysis of the phase transitions of aMCL in the context of speech separation.
\end{itemize}

\subsubsection{Separation performance on 2- and 3-speaker mixtures} \label{sec:apx:ssep_perf}

Table \ref{tab:apx:ssep_perfs} shows the average score of each separation model in the 2- and 3-speaker scenarios. 

First, we observe that PIT and aMCL achieve similar performances, both in terms of the average PIT SI-SDR and MCL SI-SDR scores, and both in the 2- and the 3- speaker settings (see lines 1 and 3 of Table \ref{tab:apx:ssep_perfs}).
Looking at the inter-seed variance, we further see that the differences in performances between these approaches are not significant. Note that MCL and aMCL have a $\mathcal{O}(mn)$ complexity, while PIT is $\mathcal{O}(m^3)$ with the Hungarian algorithm \cite{edmonds1972theoretical}. Therefore aMCL reaches the same performance as PIT with a gain in terms of complexity. Further work includes an analysis of this complexity gap when the number of speakers is high, similarly to \cite{tachibana2021towards}.

Second, we see that aMCL achieves better performance than MCL in the 2-speaker scenario, both in terms of PIT SI-SDR and MCL SI-SDR metrics (see lines 2 and 3 of Table \ref{tab:apx:ssep_perfs}). More precisely, we see that this performance discrepancy between MCL and both PIT and aMCL is due to a higher variance of the MCL training method. In fact, MCL has a higher inter-seed variance than aMCL and PIT in all settings. Indeed this method is known to be sensitive to initialization \cite{makansi2019overcoming, rupprecht2017learning}. In contrast, we see that aMCL is more robust to initialization, having a lower inter-seed variance and higher average score. This phenomenon was highlighted in Section \ref{sec:apx:theoretical_analysis}, which provided a theoretical analysis of the Euclidean case. This experiment suggests that our conclusions hold even for the source separation task, where the underlying loss function $\ell$ is non-Euclidean. 

In the subsequent Sections, we further study these two claims. In Section \ref{sec:apx:hyp_num}, we demonstrate the advantages of aMCL over PIT by letting the number of predictions $n$ exceed the number of sources $m$. In Section \ref{sec:apx:transitions}, we show that other conclusions of our theoretical analysis concerning phase transitions also seem to hold in the non-Euclidean setting. 

\begin{table}[h]
    \centering
    \caption{\textbf{Comparison of the training methods PIT, MCL, and aMCL}, for the task of source separation of 2 and 3 speakers. The performance is measured using the PIT SI-SDR metric \textit{(left)} and the MCL SI-SDR metric \textit{(right)} computed on the WSJ0-mix evaluation subset. Each method is trained using three seeds and we report inter-seed average score and standard deviation.}
    \begin{subtable}{.5\linewidth}
      \centering
        \caption{PIT SI-SDR ($\uparrow$)}
        \begin{tabular}{lll}
            \toprule        
                Method &  2 speakers  & 3 speakers\\
                \midrule
                PIT & 16.88 \hfill$\pm$ 0.10 & 10.01 \hfill$\pm$ 0.04\\
            MCL & 16.30 \hfill$\pm$ 0.59 & 10.06 \hfill$\pm$ 0.21\\
            aMCL & 16.85 \hfill$\pm$ 0.13 & 10.00 \hfill$\pm$ 0.21\\
        \bottomrule
        \end{tabular}
    \end{subtable}%
    \begin{subtable}{.5\linewidth}
      \centering
        \caption{MCL SI-SDR ($\uparrow$)}
        \begin{tabular}{lll}
            \toprule        
                Method &  2 speakers  & 3 speakers\\
                \midrule
                PIT & 16.88 \hfill$\pm$ 0.10 & 10.04 \hfill$\pm$ 0.04\\
            MCL & 16.30 \hfill$\pm$ 0.59 & 10.09 \hfill$\pm$ 0.21\\
            aMCL & 16.86 \hfill$\pm$ 0.13 & 10.04 \hfill$\pm$ 0.20\\
        \bottomrule
        \end{tabular}
    \end{subtable}   
    \label{tab:apx:ssep_perfs}
\end{table}

\subsubsection{Impact of the number of hypotheses in MCL and aMCL}\label{sec:apx:hyp_num}

\begin{figure}[h]
	\centering
	\begin{subfigure}{0.33\linewidth}
		\includegraphics[width=\linewidth]{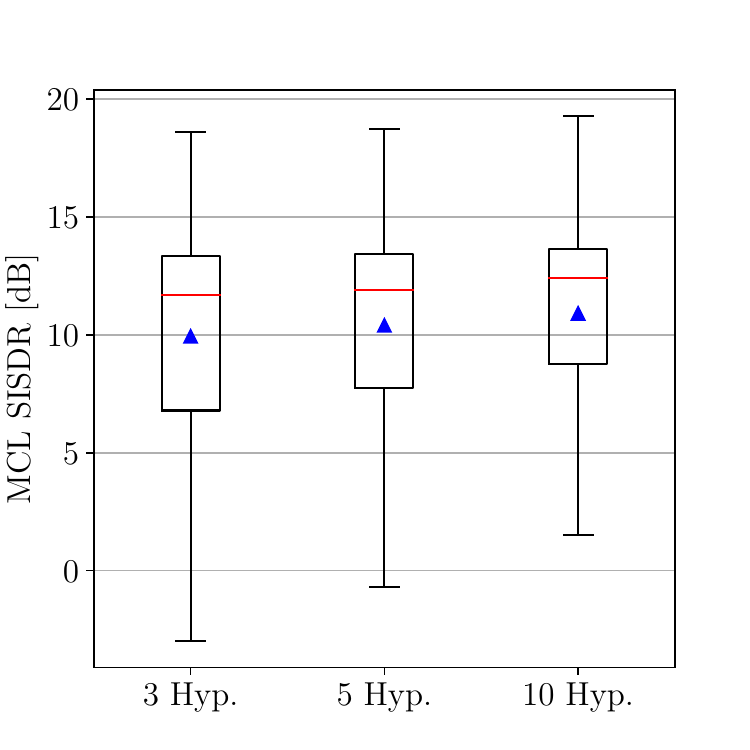}
		\caption{MCL}
		\label{fig:apx:var_hyp_a}
	\end{subfigure}
	\begin{subfigure}{0.33\linewidth}
		\includegraphics[width=\linewidth]{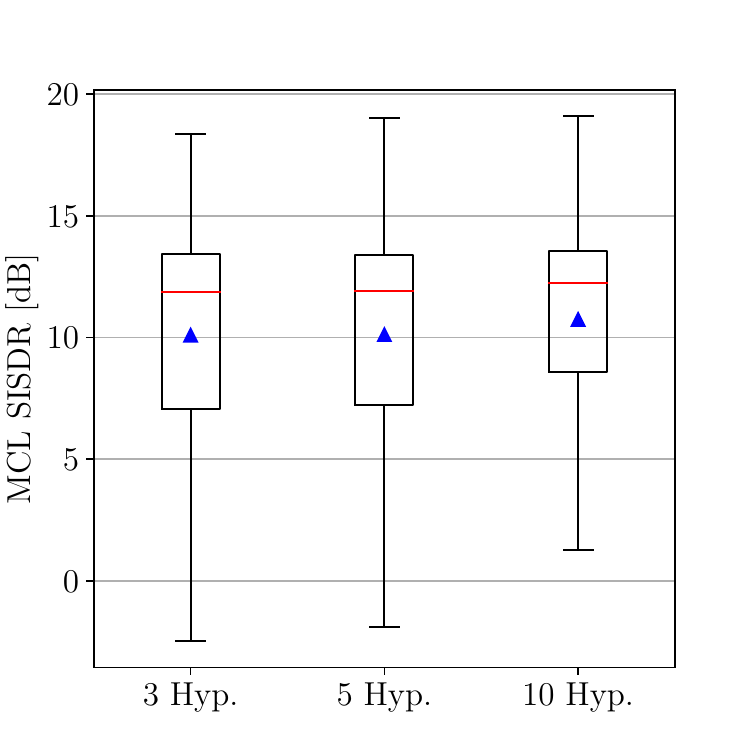}
		\caption{aMCL}
		\label{fig:apx:var_hyp_b}
	\end{subfigure}
     \caption{\textbf{Impact of the number of hypotheses.} Comparison of MCL (left) and aMCL (right) on the WSJ0-3mix dataset, using the MCL SI-SDR metric. Dataset-wise scores average (blue triangle), median (red line), and quartiles (main box) are reported. The whiskers extend the box by 1.5 times the interquartile gap. The number of hypotheses $n$ is selected among $\{3, 5, 10\}$. A higher score indicates better separation performance.}
    \label{fig:apx:var_hyp}
\end{figure}

In this Section, we study the impact of the number of hypotheses on the source separation performances. We focus on the 3-speaker scenario, and we increase the number of predictions. More precisely, we consider the cases $n\in \{3,5,10\}$. If $n=3$, the number of hypotheses is the same as the number $m$ of sources to separate. If $n\in \{5,10\}$ case the number of hypotheses is higher. Since the number of predictions and sources differ, it is not possible to compute the PIT SI-SDR metric, which is ill-defined in this setting. Therefore, we only present results for the MCL SI-SDR metric. In this experiment, we use a single seed across the considered methods to ensure a fair comparison. We report dataset-wise score statistics in Figure \ref{fig:apx:var_hyp}. 

First, we observe that the separation performance increases as the number of predictions $n$ increases, both for MCL and aMCL. This shows that the additional hypotheses are effectively used by both methods, despite the risk of collapse or convergence towards a suboptimal configuration. Recalling that MCL, aMCL, and PIT performances are similar in the case $n=m=3$, this demonstrates a strong advantage of the MCL family compared to PIT. Indeed, we can improve MCL SI-SDR score beyond what is achievable by PIT training, by increasing the number of predictions $n$. Moreover, this improvement comes at minimal computational complexity cost, as discussed in Section \ref{sec:apx:ssep_perf}. This property can be exploited to handle settings where the number of speakers is unknown. This is left to further work.

Second, we observe that there are no significant discrepancies between MCL and aMCL performances in this setting. This suggests that the main difference between these two approaches is their sensitivity to initialization.

\subsubsection{Phase transition in aMCL training}
\label{sec:apx:transitions}

\begin{figure}[b!]
	\centering
    \includegraphics[scale=0.5]{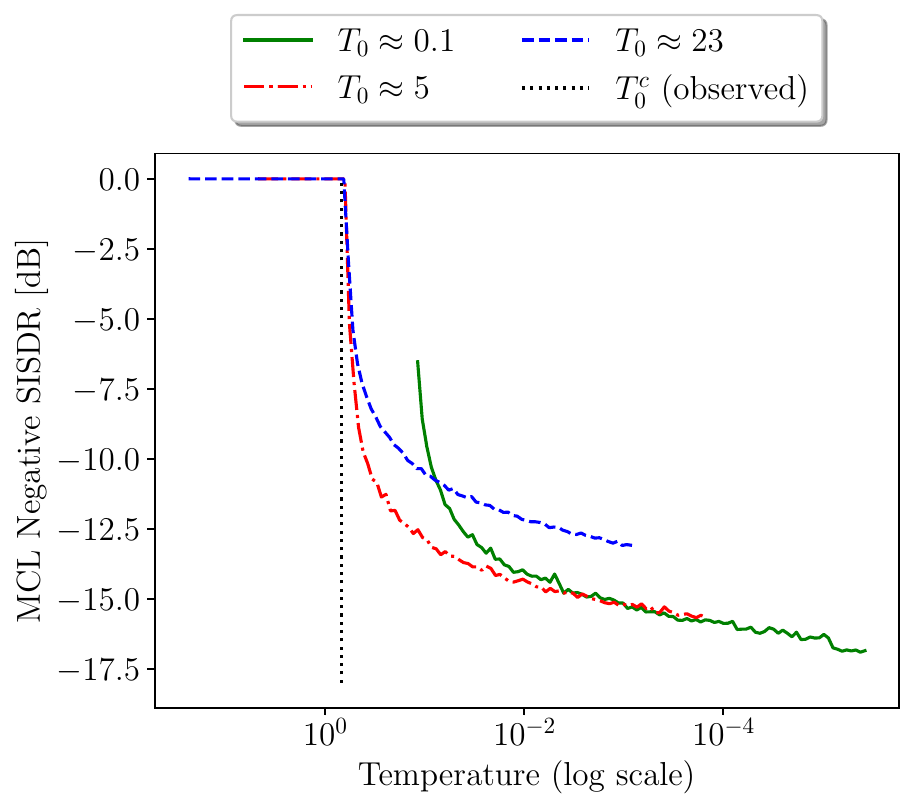}
     \caption{\textbf{Phase transition in speech separation training.} Impact of the initial temperature $T_0$ of the temperature scheduler on the source separation performance during training. The y-axis corresponds to the negative MCL SI-SDR metric. The x-axis corresponds to the temperature $T(t)$ at each training step $t$, and is displayed in logarithmic scale. Comparison of several initial temperatures $T_0\approx0.1$ (green solid line), $T_0\approx5$ (red dashed and dotted line), and $T_0\approx 23$ (blue dashed line). A lower score indicates better separation performance. }
    \label{fig:apx:transi}
\end{figure}

In this Section, we analyze the training trajectory of aMCL in speech separation. We focus on exponential schedulers defined by
\begin{equation}
    T(t) = T_0 \, \rho^t \, \bI[t < t_{\mathrm{max}}]\;,
\end{equation}
where $\rho=0.9$ is the decay factor, and $T_0\in \{0.1, 5, 23 \}$ is the initial temperature. Figure \ref{fig:apx:transi} indicates the negative MCL SI-SDR score computed during training on a validation set. In order to compare meaningfully the different temperature schedules, we plot the separation score against the temperature, regardless of the training step when this temperature is reached. 

We see that that aMCL exhibits phase transitions in this setting. Indeed, when we use a scheduler with high initial temperature $T_0\in \{5, 23\}$, the negative MCL SI-SDR metric stays on a plateau and then decreases after some critical temperature $T_0^c$ has been reached. As expected, this critical temperature seems to be independent of the initial temperature $T_0 > T_0^c$. Moreover, when using a scheduler with initial temperature $T_0\approx0.1<T_0^c$, aMCL does not undergo a phase transition. This suggests that the phase transition phenomenon exhibited in Section \ref{sec:apx:theoretical_analysis} also exists in the non-Euclidean case, even though the properties of the barycenter and the shape of the Voronoi cells have to be redefined. Further work will analyze more exhaustively the impact of the scheduler type on this phase transition phenomenon. We hypothesize that the temperature decay speed at the critical temperature $T_0^c$ will play a key role in this analysis.

\newpage

\end{document}